\pdfoutput=1
\documentclass[11pt]{article}

\usepackage[final]{acl}

\usepackage{times}
\usepackage{latexsym}
\usepackage{amsmath}
\usepackage[T1]{fontenc}
\usepackage{comment}
\usepackage{booktabs}
\usepackage{multirow}
\usepackage[utf8]{inputenc}

\usepackage{microtype}

\usepackage{inconsolata}

\usepackage{graphicx}

\title{Balcony: A Lightweight Approach to Dynamic Inference of Generative Language Models} 

\author{
    Benyamin Jamialahmadi\thanks{\ \ Equal contribution.}\textsuperscript{\rm 1},
    Parsa Kavehzadeh\footnotemark[1]\textsuperscript{\rm 1},
    Mehdi Rezagholizadeh\textsuperscript{\rm 1},
    Parsa Farinneya\textsuperscript{\rm 1}, \\
    {\bf Hossein Rajabzadeh}\textsuperscript{\rm 2},
    {\bf Aref Jafari}\textsuperscript{\rm 1}, 
    {\bf Boxing Chen}\textsuperscript{\rm 1}, and
    {\bf Marzieh S.Tahaei}\textsuperscript{\rm 1} \\
    \textsuperscript{\rm 1}Huawei Noah's Ark Lab\\
    \textsuperscript{\rm 2}University of Waterloo\\
    \{benyamin.jami76, parsareal\}@gmail.com
}

\begin{document}
\maketitle
\begin{abstract}

Deploying large language models (LLMs) in real-world applications is often hindered by strict computational and latency constraints. While dynamic inference offers the flexibility to adjust model behavior based on varying resource budgets, existing methods are frequently limited by hardware inefficiencies or performance degradation.
In this paper, we introduce \textbf{Balcony}, a simple yet highly effective framework for \textbf{depth-based dynamic inference}. By freezing the pretrained LLM and inserting additional transformer layers at selected exit points, Balcony maintains the full model’s performance while enabling real-time adaptation to different computational budgets. These additional layers are trained using a straightforward self-distillation loss, aligning the sub-model outputs with those of the full model. This approach requires significantly fewer training tokens and tunable parameters, drastically reducing computational costs compared to prior methods.
When applied to the LLaMA3-8B model, using only 0.2\% of the original pretraining data, Balcony achieves minimal performance degradation while enabling significant speedups. Remarkably, we show that Balcony outperforms state-of-the-art methods such as Flextron and Layerskip as well as other leading compression techniques on multiple models and at various scales, across a variety of benchmarks.\footnote{Please find our code and models at \url{https://github.com/benyaminjami/Balcony-LLaMA/tree/finetuning}.}

%When applied to the LLaMA3-8B model, using only 0.2\% of the original pretraining data, Balcony achieves a minimal performance drop, 2.7\% at 70\% of the full model's latency and 4.4\% at 60\% latency. Remarkably, we show that Balcony outperforms state-of-the-art methods on multiple models and other leading compression techniques across a variety of benchmarks. \mehdi{In this part, mentioning the drop of 2.7\% can be negative. We can insted just mention : Remarkably, we show that Balcony outperforms state-of-the-art methods such as Flextron and Layerskip  on multiple models at different scales and other leading compression techniques across a variety of benchmarks.  } 
\end{abstract}

\section{Introduction}
\begin{figure}[t]
\centering
  \includegraphics[width=1.0\columnwidth]{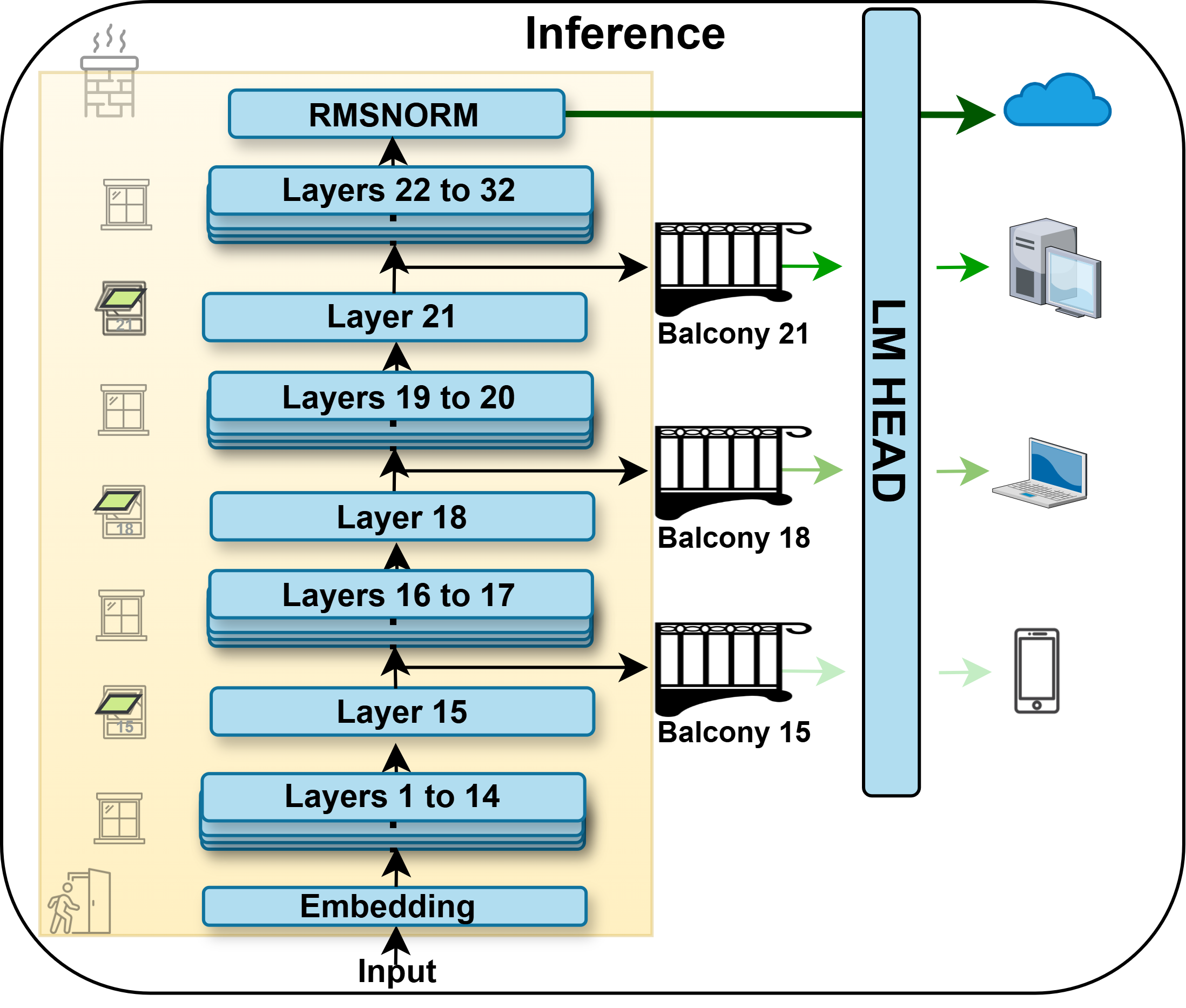}
  \caption{An overview of the \textit{Balcony} inference. Balcony preserves the base model’s performance while enabling efficient, on-the-fly adaptation to different computational budgets.}
  \label{fig:inference}
\end{figure}

% \mehdi{I had added the ulem package to this document to be able to edit using "sout" command; however, this package change the format of the references. Please comment that package before generating the final version for submission.  }
We are entering an era of rapid advancements in generative foundation models, with tens or even hundreds of billions of parameters emerging at an accelerating pace \citet{touvron2023llama,bai2023qwen,deepseekai2024deepseekllmscalingopensource}. The demand for their deployment is greater than ever. However, deploying these models in real-world applications, particularly in industrial environments, whether on edge devices or in cloud settings, is constrained by strict computational and latency requirements.

These constraints are not static; they fluctuate based on task complexity, sample difficulty, user load, and inference budget. %To overcome these limitations, dynamic inference, which allows models to adjust their computational footprint in real time, is gaining increasing significance. Such adaptability eliminates the need for costly and prohibitive retraining, enabling more efficient and scalable deployment.
To address these challenges, \textit{dynamic inference} has become increasingly important. This approach allows models to adjust their computational demands on the fly, eliminating the need for expensive and time-consuming retraining. As a result, it enables more efficient and scalable deployment. 

Dynamic inference is typically achieved in two ways: 1) width-based methods, which adjust the number of active neurons and attention heads in a model \citet{kudugunta2023matformer,flextron}, and 2) depth-based methods, which selectively reduce the number of transformer layers used during inference \citet{kavehzadeh2024sorted}.

\begin{figure}[t]
  \centering
  \includegraphics[clip, trim=0pt 2pt 2pt 0pt, width=\columnwidth]{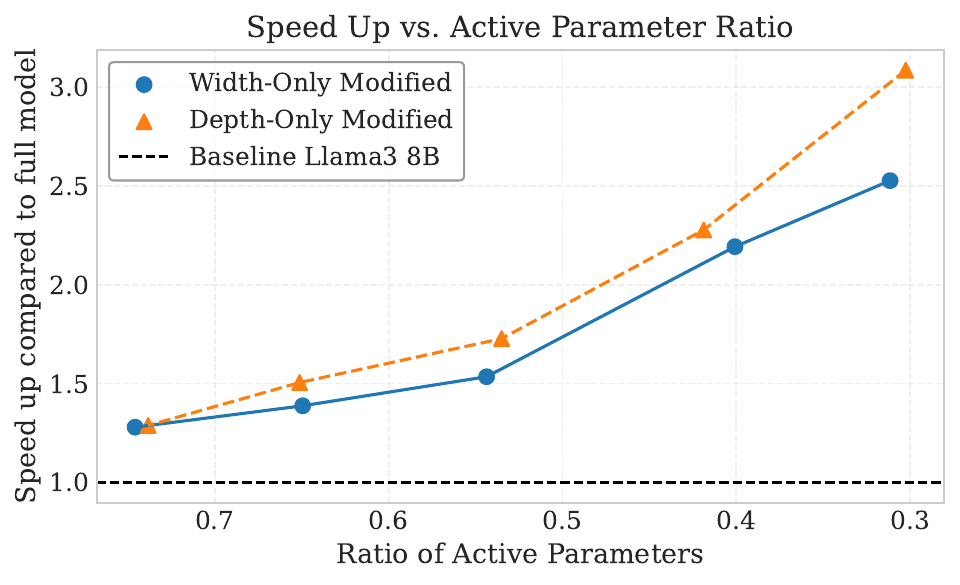}
  \caption{Speed-up as a function of the ratio of active parameters in modified versions of Llama 3 8B. Comparisons are shown between width-only and depth-only modifications, with the unmodified baseline included for reference.}
  \label{fig::depth_vs_width}
\end{figure}

However, GPUs are inherently optimized for parallelized deep computations, making depth-based compression significantly more favorable in terms of speed and efficiency. Reducing depth results in fewer sequential operations, which directly translates to lower latency and improved throughput. This effect is illustrated  in Figure \ref{fig::depth_vs_width}, where for a fixed parameter budget, reducing depth consistently yields greater speed improvements compared to reducing width.

As a result, prior research has explored methods that introduce elasticity along the depth dimension \citet{xin2020deebert,kavehzadeh2024sorted,elhoushi2024layer}. 
Despite its advantages, depth-based dynamic inference introduces a critical challenge: it often degrades both full-model accuracy and sub-model performance. This occurs because existing methods require extensive perturbation to the base model.
%l, enabling sub-models of varying computational budgets to be extracted from different exit points. 
%However, this process imposes a strong constraint on the base model, limiting its representational capacity.

Specifically, at each exit point, the sub-model must serve two competing roles: producing an intermediate representation for the next layer (to support larger sub-models) while simultaneously generating a refined output representation at the current layer. These competing objectives at the training time lead to \textit{conflicting gradients}, introducing a trade-off that ultimately compromises the accuracy of both the sub-models and the full model. \citet{rotem2023finding,kavehzadeh2024sorted}.

In this paper, we propose Balcony, a simple yet highly efficient framework for depth-based dynamic inference. By freezing the pretrained LLM and adding a decoder layer at each exit point, Balcony preserves the base model’s performance while enabling efficient, on-the-fly adaptation to different computational budgets. While there is an inherent trade-off between sub-model accuracy and computational cost, we show that adding a single transformer layer and sharing the LM head, across all sub-models achieves an optimal balance. We train the Balcony layers using a  straightforward self-distillation loss, aligning Balcony layers outputs with those of the full model.%Since the base model remains unchanged, only a few additional layers need to be uploaded, simplifying deployment.

In contrast to prior works, which also train the base model, freezing enables lossless performance on the original base model and allows for efficient tuning due to the low number of tunable parameters. Additionally, freezing facilitates seamless adaptation to different computational budgets during inference by simply switching the Balcony layers.

  %, defined as the Kullback-Leibler (KL) divergence between the outputs of Balcony exits and the base model’s final exit after the shared LLM head.

Our experiments demonstrate that for LLaMA3-8B, freezing the base model and tuning only the Balcony layers (2.5\% of full model parameters for each balcony layer), using just 0.2\% of the data compared to the 15T tokens  used for full model pretraining, yields remarkably strong results.

Notably, Balcony surpasses state-of-the-art (SoTA) dynamic inference methods, including Flextron \citet{cai2024flextronmanyinoneflexiblelarge} and LayerSkip \citet{elhoushi2024layer}, while using a minimal training approach that requires a much simpler training flow, significantly fewer training tokens and a much smaller number of tunable parameters (see Related Work for details on the training strategies used in these methods).
This paper makes the following key contributions:
\begin{itemize}
    \item Introducing Balcony, a depth-based dynamic inference framework that employs single transformer layers at exit points while freezing the base model.  
    \vspace{-2pt}
    \item Efficient tuning of Balcony through a self-distillation loss on a small dataset, significantly reducing training costs compared to prior methods while outperforming Flextron, LayerSkip and SoTA compression methods.  
    \vspace{-2pt}
   \item An extensive evaluation of the proposed framework through ablation studies on various components of the method, including pretraining on a 1B-parameter LLM.
\end{itemize}
%Recently, \citet{kudugunta2023matformer} and \citet{cai2024flextronmanyinoneflexiblelarge} proposed methods for creating a single, customizable model with multiple sub-networks tailored to different resource budgets. 

%Moreover, in the proposed approaches, there is no mechanism to distinguish between the role of an intermediate layer—whether it functions as a normal intermediate layer or as an exit. This lack of differentiation impacts the performance of the sub-models. 

\section{Related work}

%\subsection{Dynamic inferece for generative LLMs}

Dynamic inference has gained significant attention over the past decade, with various methods proposed to make CNNs and small-scale encoder based NLP models dynamic \citet{yu2018slimmable, yu2019universally,li2021dynamic,cai2020onceforalltrainnetworkspecialize,xin2020deebert,hou2020dynabert,kusupati2022matryoshka}.
These approaches often require sophisticated and heavy training, are architecture-dependent  and hence have not yet been effectively translated to modern large-scale generative LLMs. This paper focuses specifically on generative LLMs, emphasizing approaches applied within the realm of generative AI.

MatFormer \citet{kudugunta2023matformer} introduces a nested architecture along the width that enables the extraction of multiple sub-models by incorporating a nested Feed Forward Network (FFN) block structure from a single trained network. The largest reported MatFormer model is an 850M decoder-only language model (MatLM), from which smaller models ranging from 582M to 850M parameters can be derived. MatFormer demonstrates superior performance compared to independently trained models and older methods like OFA \citet{cai2020onceforalltrainnetworkspecialize} and DynaBERT \citet{hou2020dynabert}.%, its results are inferior to Flextron, as shown in the Flextron paper. 

Building on this, Flextron proposed a more sophisticated approach that integrates a nested elastic structure with input-adaptive routing, allowing automatic token processing through sub-networks. However, their training strategy is highly sophisticated: first, they train a large number of submodels using their nested architecture. Then, they train routers to select the appropriate submodels based on a given budget (for static inference) or dynamically for each token (for adaptive inference). However, training the router is challenging due to limited gradient flow. To address this, they introduce an auxiliary model that provides the necessary signals to facilitate the router’s training. We demonstrate that, with significantly fewer training tokens and a far simpler training approach, our submodels achieve superior performance.

Similar to Balcony, SortedLLaMA \citet{kavehzadeh2024sorted} explores elasticity along the depth dimension by extending the SortedNet \citet{valipour2023sortednet} training technique to generative LLMs. They eliminate the need for pretraining by replacing standard fine-tuning with sorted fine-tuning. LayerSkip \citet{elhoushi2024layer} is another recent dynamic depth approach that is used along speculative decoding for faster inference. During training, the method employs layer dropout with increasing rates for deeper layers and applies an early exit loss on all transformer layers while sharing the LM head.

Note that the approaches mentioned above perturb the base model to create a nested design of submodels, allowing multiple submodels to be hosted within a shared architecture while reducing memory overhead. This, in turn, leads to performance degradation in the full model. In contrast, our method achieves the same objective without enforcing a nested structure. Instead, we freeze the base model and train only the Balcony Exit and auxiliary tokens, ensuring that the base model's performance remains intact and uncompromised by nested biases. Additionally, since all submodels share the same base model, the memory overhead of loading multiple submodels simultaneously or switching between them remains manageable.

\begin{figure}[t]
\centering
\includegraphics[ width=0.9\columnwidth]{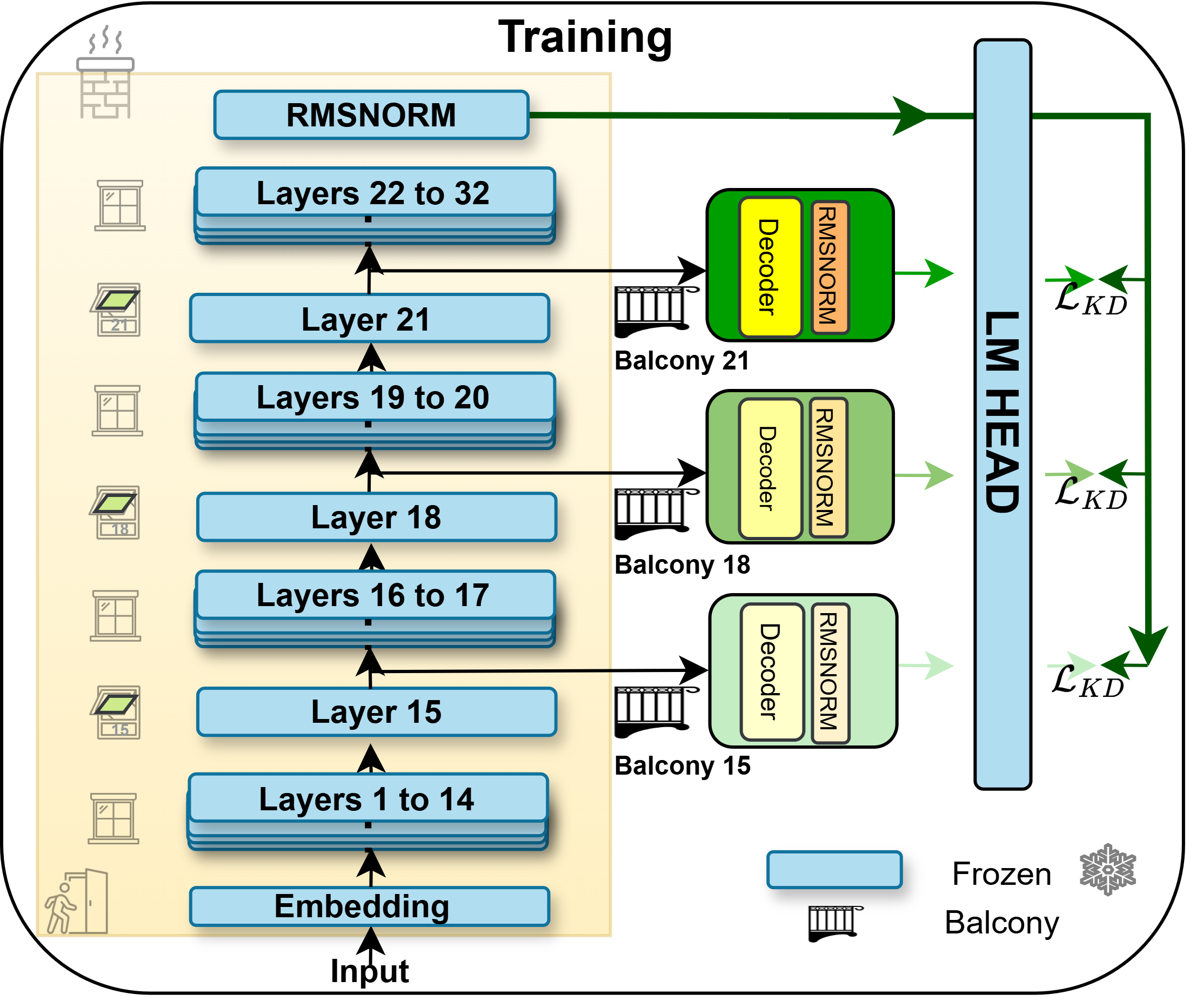}
\caption{ Training in the Balcony framework: By freezing the pretrained base LLM and adding a decoder layer at each exit point, Balcony can outperform SoTA with significantly fewer training tokens.}
  \label{fig:experiments}
\end{figure}

\vspace{-2pt}
\section{Balcony framework}
In this section, we present the architecture, training, and inference methods for the proposed framework.
%\subsection{Balcony Exit}
Consider a model \( M \) with \( N \) layers, where each layer is represented by the function \( X_{i} = f(X_{i-1}, W_i) \), with \( i \in [1, N] \) indexing the layers, \( X_{i-1} \) being the input to layer \( i \) (with dimensions \( B \times S \times D \) for batch size, sequence length, and embedding dimension), and \( W_i \) denoting the parameters for that layer. In this context, the term "layers" includes not only traditional transformer blocks but also other depthwise modules, such as Mamba~\citet{gu2023mamba} and Mixture of Experts (MoE)~\citet{fedus2022switch} blocks. 
 
Our objective is to make the model \( M \) dynamic in depth, allowing it to adapt to user-defined goals, such as latency, memory, and accuracy. To achieve this, we introduce exit points based on the desired inference time and resource budget. For instance, in cloud computing, we can select exit layers depending on the query-per-second rate. Similarly, on edge devices, one can extract submodels depending on the available computation budget.

An \textbf{exit point} in our framework refers to a designated layer within the model where inference can be halted, allowing a prediction to be generated using a lightweight \textit{Balcony module} instead of processing all layers of the model. 
We introduce a set of \textbf{exit points} \( \mathcal{E} \subseteq \{1, 2, \dots, N\} \), each associated with a \textit{Balcony module}. At each exit point \( j \in \mathcal{E} \), the intermediate representation \( X_{j} \) is processed by a Balcony module, defined as:
\begin{equation}
    \mathcal{X'}_{j} = f_b(X_{j}, \mathcal{W'}_j),
\end{equation}
where \( f_b \) represents a Balcony module composed of a decoder layer(e.g a transformer block) followed by a normalization layer, \( \mathcal{W'}_j \) is the set of parameters for the Balcony module, and \( \mathcal{X}'_j \) is the output of the Balcony module at exit point \( j \).

% For a 32-layer model suited for cloud deployment, we might choose the entire model, while we can add extra exit points at layers 16, 20, 28 for edge and laptop and desktop deployment, respectively. Specifically, exit points are defined as \( \text{Exit Points} = \{\text{Layer}_{16}, \text{Layer}_{20}, \text{Layer}_{28}\}\).

% As mentioned earlier, to address the conflicting gradient issue inherent in early-exit methods, we freeze the base model \( M \) and introduce a transformer layer at each exit point. We refer to these as \emph{Balcony} layers: \( \text{Balconies} = \{\text{Balcony}_16, \text{Balcony}_20, \text{Balcony}_{28}\} \).

 %In our experiments our models follow the LLaMA architecture, and hence transformer layers incorporate Grouped-Query Attention (GQA) for efficiency, Rotary Positional Embeddings (RoPE) for long-context modeling, and a SwiGLU-activated feedforward network for improved expressivity. 

% The attention mechanism computes:
% \begin{equation}
% A = \text{softmax} \left( \frac{Q K^T}{\sqrt{d_k}} \right) V
% \end{equation}

% where RoPE is applied to \( Q \) and \( K \) before computing attention. The feedforward network (FFN) follows:
% \begin{equation}
% \text{FFN}(X) = \sigma(X W_1) \odot (X W_2) W_3
% \end{equation}
% with SwiGLU activation replacing standard ReLU/GELU. Each layer applies pre-normalization (RMSNorm) before attention and FFN, with residual connections ensuring stable gradient flow. 
At each exit point, the intermediate output \( X_{j} \) is forwarded to the corresponding Balcony module $f(X_{j}, \mathcal{W'}_j)$ then passes through an RMSNorm layer, followed by a shared LM head and a softmax function to produce the final probabilities. The LM head is the same as that of the original model and is shared across all submodels. The transformer layer in Balcony uses the same architecture as the base model. %\mehdi{We could make our method a bit more general by introducing a new function g instead of f and then say that in this work we consider g to follow f.  }

During training, self-distillation is used to align the probability distribution \( p(\cdot;  W_{1:j}, \mathcal{W}'_j,) \) of each submodel with the full model's distribution \( p(\cdot; W_{1:N}) \).
This is achieved by minimizing the Kullback-Leibler (KL) divergence across all possible next tokens at position \( t \). The objective function is given by:
\begin{equation}
\mathcal{L}= \sum_{j \in \mathcal{E}} \text{KL}\left( p(\cdot; W_{1:N}) \parallel p(\cdot; W_{1:j}, \mathcal{W}'_j) \right),
\end{equation}
where \( \mathcal{W}'_j \) represents the balcony layer inserted after layer \( j \), and \( \text{KL}(\cdot \parallel \cdot) \) denotes the Kullback-Leibler divergence. %The set \({\mathcal{E}}\) refers to the set of balcony exit layers.

Note that both the base model parameters $W_{1:N}$ and the LM head remain frozen during training. 
In our framework, Balcony layers are initialized from the last transformer layer of the trained model (see Section \ref{sec:ablation} for the impact of this initialization). The rationale behind this choice is that the last layer is already aligned with the shared LM head, which helps in seamlessly integrating the Balcony layers for effective submodel extraction.

Since the base model remains frozen, the training of Balcony layers is independent; each Balcony layer receives gradients based only on its corresponding submodel’s loss. Consequently, while all Balcony layers can be trained in a single training round to minimize computation, they can also be added and trained individually  without impacting any other submodel. %This flexibility allows for scalable dynamic inference, enabling efficient adaptation to different computational budgets.

\begin{comment}
\subsection{Condition/reference/auxiliary Tokens}
Another challenge in depth-wise dynamic models is that certain intermediate layers have two conflicting roles: serving as an intermediate representation for the next layer or acting as a layer directly followed by an exit. These roles require different levels of abstraction. One solution to address this is to encode this info \textbf{auxiliary tokens} that are learned and unique to each submodel. Thus through these auxilary token will condition a alyer to prepare information for the balcony e

During inference, the auxiliary tokens remain fixed and hence their computations can be carried out offline acting as a conditioning for a given submodel.

\end{comment}
\begin{table*}[htb]
\centering
\caption{We evaluate the downstream task performance of Balcony, comparing it against Flextron, LayerSkip, open-source models, and other compression methods. For LayerSkip, we evaluated their publicly available models. For all other baselines, the results are taken from their papers. We report zero-shot accuracy on ARC-easy, LAMBADA, PIQA, and WinoGrande, along with 5-shot performance on MMLU. For LayerSkip, we evaluated their publicly available models. \#Params denotes the number of non-embedding parameters.}
\label{table:downstream_performace}
\resizebox{\textwidth}{!}{
\begin{tabular}{l| l| c |c c c c c c }
\toprule

\textbf{Method} & \textbf{Model}  & \textbf{\# Params} & \textbf{ARC-E} & \textbf{LAMBADA} & \textbf{PIQA} & \textbf{Winogrande} & \textbf{MMLU (5)}  & \textbf{Avg. (Drop) } \\
\midrule

\multirow{1}{*}{\rotatebox{0}{\textbf{Base}}}
& \textbf{Llama2-7B-Full model} & 6.5B & 76.3 &	71.1& 78.1& 69.1& 45.9&	68.1 \\
\midrule

\multirow{4}{*}{\rotatebox{90}{\textbf{Balcony}}}
&\textbf{Balcony-XL}  & 6.5B & 76.3 &	71.1 & 78.1 & 69.1 & 45.9 &	68.1(\textbf{0}) \\
&\textbf{Balcony-L} & 4.4B & 72 & 67 & 75.9 & 67.5 & 45 & 65.5  (\textbf{-2.6}) \\
&\textbf{Balcony-M} & 3.8B & 68.9 & 61.3 & 75.2 & 66 & 43 & 62.9  (\textbf{-5.2}) \\
&\textbf{Balcony-S} & 3.2B & 64.9 & 54.9 & 73.5 & 63.8 & 39.8 & 59.4  (-8.7) \\

\midrule

\multirow{4}{*}{\rotatebox{90}{\textbf{LayerSkip}}}

& \textbf{Layerskip-XL} & 6.5B & 76.5 & 70.5 & 77.6 & 70.3 & 43.2 & 67.6 (-0.5) \\
&\textbf{Layerskip-L} & 4.4B & 68.5 & 65.9 & 73.7 & 66.4 & 42.4 & 63.4 (-4.7) \\
&\textbf{Layerskip-M} & 3.8B & 61.4 & 55.1 & 71.1 & 65.8 & 42.4 & 59.2 (-8.9) \\
&\textbf{Layerskip-S} & 3.2B & 50.3 & 43.5 & 68.6 & 63.9 & 37.8 & 52.8 (-15.3) \\

\midrule
\multirow{4}{*}{\rotatebox{90}{\textbf{Flextron}}}
&\textbf{Full} & 6.5B & 75.1 & 71.5 & 77.5 & 69.1 & 45.1 & 67.7 (-0.4) \\
&\textbf{Dynamic 7x} & 4.1B & 68.6 & 65.1 & 76.1 & 63.7 & 42.2 & 63.1 (-5) \\
&\textbf{Dynamic 6x} & 3.9B & 67.1 & 63.8 & 74.9 & 62.2 & 39.4 & 61.5 (-6.6) \\
&\textbf{Dynamic 5x} & 3.4B & 66.5 & 62.9 & 74.1 & 62 & 36.8 & 60.5 (\textbf{-7.6}) \\

%  & Static-0.7$\times$ & 4.2 B & 65.8 & 64.2 & 75.6 & 62.3 & 41.9 & \\
%  & Dynamic-0.7$\times$  & 4.1 B & 68.6  & 65.1 & 76.1 & 63.7 & 42.2 & \\
%  \cmidrule{2-9}
%  & Static-0.6$\times$ & 4.0 B & 66.1 & 63.8 & 75.0 & 62.1 & 37.7 & \\
%  & Dynamic-0.6$\times$  & 3.9B  & 67.1 & 63.8 & 74.9 & 62.2 & 39.4 & \\
%  \cmidrule{2-9}
%  & Static-0.5$\times$ & 3.5 B & 65.9 & 61.7 & 74.8 & 61.9 & 35.9 & \\
% & Dynamic-0.5$\times$  & 3.4 B & 66.5 & 62.9 & 74.1 & 62.0 & 36.8 & \\
\midrule

\midrule
\midrule

\multirow{1}{*}{\rotatebox{0}{\textbf{Base}}}
& \textbf{LLaMA3-8B}  
  & 6.9B & 81.8 & 71.2 & 80.1 & 73.6 & 65.0 & 74.3 \\
 \midrule
\multirow{4}{*}{\rotatebox{90}{\textbf{Balcony}}}
 & \textbf{Balcony-XL} 
 & 6.9B & 81.8 & 71.2 & 80.1 & 73.6 & 65.0 & 74.3 (\textbf{0}) \\
% & \textbf{Balcony-L} 
%   & 6.3B & 80.3 & 69.4 & 79.4 & 73.6 & 64.5 & 73.4  (\textbf{-0.9})\\
 % & \textbf{Balcony-LM} 
   % & 5.4B & 79.5 & 68.0 & 78.4 & 72.8 & 62.2 & 72.2 (\textbf{-2.1})\\
 & \textbf{Balcony-L} 
   & 4.7B & 77.0 & 67.0 & 77.4 & 72.3 & 64.0 & 71.6 (\textbf{-2.7})\\
   
 & \textbf{Balcony-M} 
   % & 4.6B & 77.9 & 65.2 & 78.0 & 72.3 & 64.1 & 71.5  (\textbf{-2.8})\\
   & 4.4B & 75.7 & 62.7 & 76.7 & 69.7 & 64.4 & 69.9 (\textbf{-4.4})\\
   
 & \textbf{Balcony-S} 
 & 3.4B & 70.0 & 54.5 & 	75.0 & 	68.1 & 	48.3 & 63.2 (\textbf{-11.1})\\
 
 % & 3.7B & 72.1 & 			56.3 & 	75.2& 	68.2 & 	48.8 & 	51.8	64.1 (\textbf{-10.2})\\

% & \textbf{Balcony-S} 
%    & 3.7 B & 72.1 & 56.3 & 75.2 & 68.2 & 48.8 & 64.1 (-10.2)\\
\midrule

\multirow{4}{*}{\rotatebox{90}{\textbf{LayerSkip}}}
& \textbf{LayerSkip-XL }& 6.9B & 79.7 & 72.5 & 80.1 & 73.8 & 59.6 & 73.2 (-1.1)\\ 
%\cmidrule{2-9}
% & Static-0.7$\times$ &  4.1 B & 66.7 & 62.9 & 75.1 & 63.9 & 28.7 &  \\
 & \textbf{LayerSkip-L}  & 4.7B  & 73.2 & 61.8 & 77.2 & 71.0 & 59.1 & 68.5 (-5.8)\\ 
 %\cmidrule{2-9}
 %& Static-0.6$\times$  & 3.9B  &  66.2 & 62.8 & 75.6 & 62.7 & 28.8 & \\ 
 & \textbf{LayerSkip-M}  & 4.4B &   68.6 & 60.8 & 74.2 & 70.1 & 59.3  & 66.7 (-7.6) \\ 
 %\cmidrule{2-9}
% & Static-0.5$\times$  & 3.4B &  64.2 & 62.0 & 74.9 & 61.7 & 25.1 & \\
& \textbf{LayerSkip-S} & 3.4B & 59.0 & 47.7 & 70.4 & 66.4 & 37.8 &  56.3 (-18)\\
\midrule
% \midrule
% \midrule

\multirow{4}{*}{\rotatebox{90}{\textbf{Flextron}}}
& \textbf{Full-Flextron-8B} & 6.4B & 71.7 & 69.7 & 79.4 &	68.8 & 35.4 & 65\\ 
%\cmidrule{2-9}
% & Static-0.7$\times$ &  4.1B & 66.7 & 62.9 & 75.1 & 63.9 & 28.7 &  \\
 & \textbf{Dynamic-0.7$\times$}  & 4.3B  & 67.0 & 64.8 & 75.9 & 64.1 & 30.0 & 60.4\\ 
 %\cmidrule{2-9}
 %& Static-0.6$\times$  & 3.9B  &  66.2 & 62.8 & 75.6 & 62.7 & 28.8 & \\ 
 & \textbf{Dynamic-0.6$\times$}  & 3.9B &   66.2 & 63.7 & 76.1 & 62.7 & 29.1  & 59.6 \\ 
 %\cmidrule{2-9}
% & Static-0.5$\times$  & 3.4B &  64.2 & 62.0 & 74.9 & 61.7 & 25.1 & \\
& \textbf{Dynamic-0.5$\times$} & 3.3B & 65.0 & 62.5 & 75.8 & 61.8 & 27.1 &  58.4\\
\midrule

\midrule
\midrule

\multirow{5}{*}{\rotatebox{90}{\textbf{Open-Source}}} 
% & Llama2-7B & 6.5B & 75.2  & 68.2 & 78.8 & 69.2 & 45.3 &  \\ 
 &\textbf{OpenLLaMA-7Bv2} & 6.5B & 69.5 & 63.8 & 79.9 & 66.0 & 40.4 & 63.92 \\
 &\textbf{OpenLLaMA-3Bv2} & 3.2B & 63.7 & 59.1 & 78.1 & 63.3 &25.7 & 58.0 \\

%& GPT3-8B & 6.4B & 70.1 & 70.5 & 79.7 & 69.8& 40.2 &  \\
% & Pythia-1.4B & 1.2B & 53.9 & 46.8 & 70.6 & 57.1 & 25.6 & \\
% & Pythia-2.8B & 2.5B & 57.9 & 50.1 & 73.8 & 58.6 & 26.8 & \\ 
%& Pythia-6.9B & 6.4B  & 60.2 & 47.1 & 75.2 & 59.9 & 25.5 & \\ 

%& Sheared-LLaMA-1.3B & 1.2B & 61.5 & 61.0 & 73.4 & 57.9 & 25.7 & \\
%& Sheared-LLaMA-2.7B & 2.5B & 67.0 & 68.4 & 75.8 & 64.2 & 26.4 & \\ 
& \textbf{NutePrune} & 3.2B & 51.7 & - & 71.0 & 57.5 & - & -\\
%& \textbf{LLM-Pruner} & 4.5B & 59.2 & - & 73.4 & 64.2 & 23.9 &  - \\
& \textbf{Compresso-compressed LLaMA-7B} & 4.5B & 66.0 &  - & 72.9 & 63.4 & 25.9 & - \\
& \textbf{LaCo-compressed LLaMA2-7B }& 4.7B & - & - & 69.8 & - & 26.5 &  -  \\
%& \textbf{SliceGPT} & 4.8B & - & - & 66.2 & - & 28.9 &  - \\

\bottomrule

\end{tabular}

}
\end{table*}

\begin{table*}[h!]
\centering
\caption{Training cost comparison for Flextron, LayerSkip and Balcony methods applied to LLaMA2-7B and LLaMA3-8B models, with costs presented in terms of tokens. In the Balcony method, the base model is frozen, and only the Balcony layers are updated. Each Balcony layer is a single transformer layer, comprising 202M parameters. The training cost for Flextron is taken from the paper \citet{flextron}. Additionally, the table includes the percentage of pretraining cost relative to the total pretraining cost for each method.}
\label{table:pretraining_cost}
\resizebox{0.9\textwidth}{!}{
\begin{tabular}{l|l|c c c}
\midrule
 & \textbf{Method} & \textbf{Number of tunable parameters} & \textbf{Training cost in tokens} & \textbf{Percentage of pretraining tokens} \\
\midrule
\multirow{3}{*}{\rotatebox{0}{\textbf{LLaMA2-7B}}} 
& \textbf{Pretraining} & 7B & 2T & 100\%\\ 

& \textbf{Balcony}  &  $3\times200\text{M} = \textbf{600M}$  & \textbf{31B} & \textbf{1.5\%} \\
    & \textbf{Flextron}    &  7B  & 89.9B (Main model excluding the router) %) 1.049B + 2.62B(router)
    & {4.50\%} % + {0.18\%} 
    \\
     & \textbf{LayerSkip}   &   7B & 52B & 2.6\%\\
\midrule
\multirow{2}{*}{\rotatebox{0}{\textbf{LLaMA3-8B}}}
  & \textbf{Pretraining} &  8B  & 15T & 100\%  \\
  & \textbf{Balcony}     &  $3\times200\text{M} = \textbf{600M}$  & \textbf{31B} & \textbf{0.2\%}\\
   & \textbf{LayerSkip}     &  8B  & 419B & 2.8\% \\
\hline
\end{tabular}
}
\end{table*}

\vspace{-2pt}
\section{Experiments}
\subsection{Setup} \label{sec:experiments_setup}
\noindent \textbf{Model Configuration}
We compare the performance of Balcony to SoTA methods using two models, LLaMA3-8B and LLaMA2-7B. The rationale behind selecting these models is that prior SoTA works have reported their results on them. To explore the methods at a smaller scale, we use an LLM with the same architecture as LLaMA3-1B and train it on 
FineWebEDU and Cosmopedia V2, which are part of the SmoLLM 
corpus \citet{benallal2024smollmcorpus} from Hugging Face. We refer to this model as LLM-1B. This model provides a a baseline on which Balcony and sortedNet method are applied to. It also provides  a baseline for our training from scratch experiment.

\smallskip \noindent \textbf{Training Details} For fine-tuning, both LLaMA3-8B and LLaMA2-7B were trained with a batch size of 256 for 30K steps using a cosine learning rate scheduler with a maximum learning rate of $5e^{-4}$. The sequence length was set to 4,096 tokens, and the training corpus consisted of 31.5B tokens from Cosmopedia V2. LLM-1B followed the same fine-tuning setup but with a sequence length of 2,048 tokens and 15.7B tokens from Cosmopedia V2. For pretraining, LLM-1B was trained from scratch using FineWebEDU and Cosmopedia V2, with a batch size of 384 for 500K steps, a sequence length of 2,048 tokens, and a learning rate of $5e^{-4}$ following a trapezoidal scheduler. For pretraining the LLM-1B model from scratch, we used 384B tokens sourced from FineWebEDU and Cosmopedia V2.

%  Both LLaMA3-8B and LLaMA2-7B models were fine-tuned with a sequence length of 4,048 tokens. This tuning process utilized 31B tokens, upsampled from the Cosmopedia V2 dataset. For tuning LLM-1B, we used 15.7B tokens from Cosmopedia V2 while maintaining the sequence length of 2,048 tokens.

% The batch size and number of iterations for continued training of all models were set to 256 and 3K iterations, respectively. For pretraining the LLM-1B model from scratch, we used 384B tokens sourced from FineWebEDU and Cosmopedia V2.

\smallskip\noindent \textbf{Baselines}
For the baseline in this paper, Flextron is used as the SoTA method in width-based dynamic inference, and LayerSkip and Sorted are considered for depth-based dynamic inference. In the comparison, the number of non-embedding parameters is reported. Regarding speedup, for depth-based methods, similar speedup can be achieved across different methods with the same number of parameters. However, for width-based methods, the same number of parameters results in lower speedup, as shown in Figure \ref{fig::depth_vs_width}. Since Flextron is not open-sourced, speedup comparisons cannot be reported. Nonetheless, for the same number of parameters, Balcony is expected to deliver better speedup. 
Furthermore, we contrast our method with several prominent open-source and compression model families, specifically, OpenLLaMA~\citet{openlm2023openllama}, Compresso~\citet{guo2023compresso}, NutePruner \citet{li2024nuteprune}, SliceGPT~\citet{ashkboos2024slicegpt}, and LaCo~\citet{yang2024laco}. 
Evaluation is performed on ARC \cite{clark2018think}, BoolQ \cite{boolq}, OpenbookQA \cite{openbookqa}, PIQA \cite{Bisk2020}, WinoGrande \cite{sakaguchi2021winogrande}, LAMBADA \cite{lambada}, 5-shot MMLU \cite{hendrycks2020measuring}, and 10-shot HellaSwag \cite{zellers2019hellaswag}. These evaluations were conducted using the LM-Evaluation-Harness repository \cite{eval-harness}.

\vspace{-2pt}
\subsection{Results}
\smallskip \noindent \textbf{Analysis of speedup in dynamic depth vs dynamic width} % To motivate the use of depth-based model pruning in our Balcony framework, we conducted an empirical evaluation comparing the impact of width pruning and depth pruning on model latency. Our results, summarized in Figure~\ref{fig::depth_vs_width} provide clear evidence that depth pruning is a more effective strategy for reducing inference latency Ll.
To evaluate the effectiveness of depth-based model pruning in our Balcony framework, we conducted an empirical analysis comparing the impact of width pruning and depth pruning on model latency.% Our findings, summarized in Figure~\ref{fig::depth_vs_width}, demonstrate that depth pruning provides a more substantial reduction in inference latency.
The latency measurements in Figure~\ref{fig::depth_vs_width} were obtained using vLLM~\citet{kwon2023efficient} for efficient deployment and were performed on an NVIDIA V100 32GB GPU, prompt size of 32, output size of 2048. These measurements compare both depth and width pruning on LlaMA38B. In depth pruning, the number of hidden layers was reduced from 32 layers to fewer layers, leading to an almost linear reduction in the number of active parameters proportional to the number of layers. For width pruning, we initially reduced the intermediate size of the MLP block until it reached the hidden size, followed by reducing the number of attention heads. 

Figure~\ref{fig::depth_vs_width} clearly illustrates that for all tested parameter ratios, representing pruned models relative to the non-pruned base model, depth pruning consistently yields higher speed-ups compared to width pruning. This reinforces the effectiveness of depth-based pruning strategies in achieving significant latency reductions. % while maintaining model integrity.

\smallskip\noindent \textbf{Balcony performance} %In this section, we asses the performance of Balcony on several downstream tasks, as presented in Table~\ref{table:downstream_performace}. These tasks include ARC-easy~\citet{clark2018think}, PIQA~\citet{Bisk2020}, WinoGrande~\citet{sakaguchi2021winogrande}, and MMLU~\citet{hendrycks2020measuring}. We report 5-shot performance for MMLU, while zero-shot performance is reported for the other tasks.
In this section, we assess Balcony's performance on several downstream tasks, as shown in Table~\ref{table:downstream_performace}. We use LLaMA2-7B as the baseline and compare it to Flextron and LayerSkip. The results for Flextron are taken from their original paper, while for LayerSkip, the dynamic models are open-sourced, so we conducted our own evaluation using the provided dynamic LLaMA-7B. For Flextron, we report the dynamic version, as it demonstrates superior performance compared to the static version. 
Among the three methods, Balcony is the only one that maintains the performance of the full model by freezing it during tuning. In contrast, both Flextron and LayerSkip experience a performance drop for their full models. For smaller submodels, with reductions of approximately 7x, 6x, and 5x in the number of non-embedding parameters, Balcony shows a significantly smaller performance drop compared to the baselines.
The only exception is Balcony-S, which, with 3.2B parameters, experiences a 1.1\% larger drop than the Dynamic Flextron model at 3.4B parameters.

The same evaluation is also performed on LLaMA-3-8B. Here, we compare the base model to those of Balcony and LayerSkip. Similarly, the LayerSkip results are obtained using their open-sourced model. For Flextron, their 8B model (Flextron-8B) does not come from the same base model and is therefore placed in a separate section of the paper. It can be observed that for all submodel sizes, from Small to XL, Balcony provides significantly less performance drop than LayerSkip across all submodels.

\begin{figure}[h!]
\centering
  \includegraphics[clip, trim=1pt 1pt 1pt 1pt, width=0.85\columnwidth]{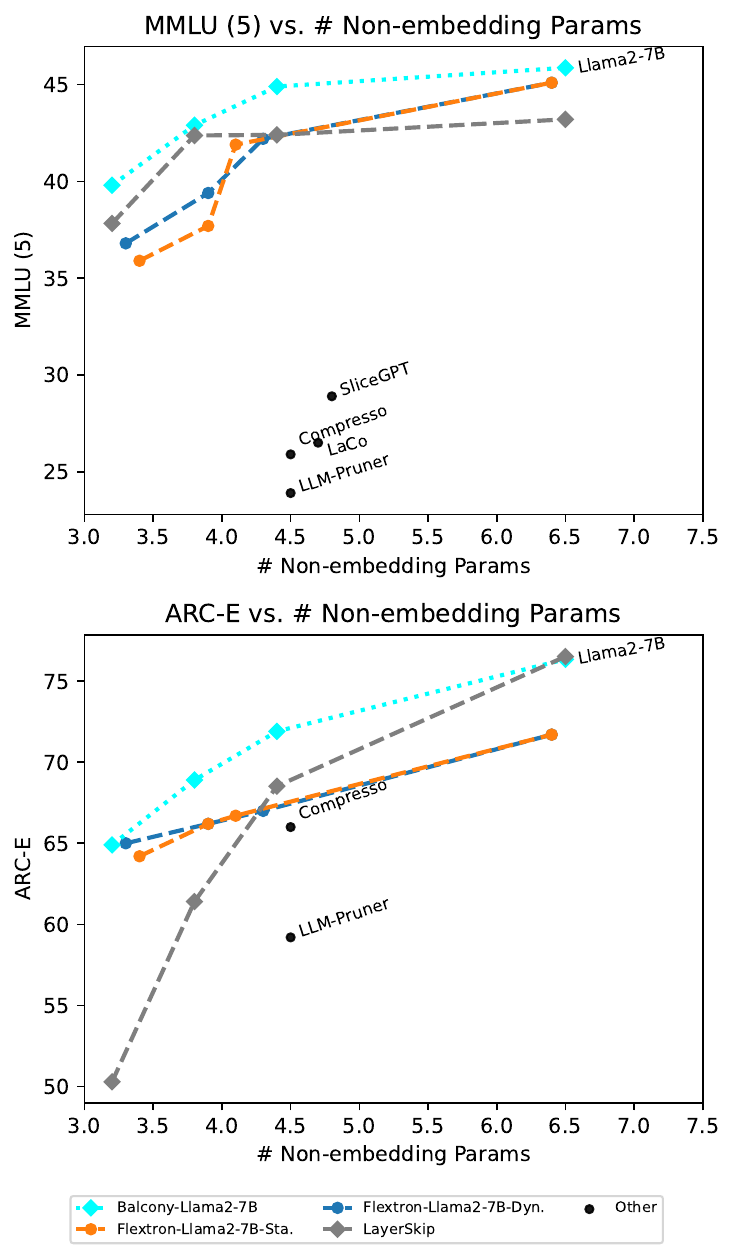}
  \caption{The Balcony-LLaMA2-7B model family demonstrates superior performance on MMLU and ARC-E compared to Flextron-Dynamic/Static, LayerSkip and post-hoc compression methods like Compresso, LLM-Pruner, SliceGPT, and LaCo.}

  \label{fig:performance_plot}
\end{figure}

Figure \ref{fig:performance_plot} plots the trade-off between accuracy and the number of parameters for the Balcony-LLaMA7B family models and compares them against those of Flextron-Dynamic, Flextron-static, and LayerSkip, as well as post-hoc compression methods like Compresso, LLM-Pruner, SliceGPT, and LaCo. The Balcony model family achieves superior performance on both MMLU and ARC-E compared to all the baselines.

\smallskip\noindent \textbf{Balcony speedup}
Table \ref{tab:balcony-latency} shows the latency of Balcony family models.The latency measurements  in seconds are based on vLLM \citet{kwon2023efficient} for efficient deployment and conducted on an NVIDIA V100 32GB GPU. The latency is measured in seconds for 30 dummy input samples with a prompting length of 32, a generation length of 2048, a batch size of 2 and float16 precision.

\begin{table}[h]
\centering
\caption{Average latency (in seconds) for Balcony family models at different scaling factors. Speed-up factors (relative to the full model) are shown in parentheses.}
\label{tab:balcony-latency}
\resizebox{\columnwidth}{!}{
\begin{tabular}{lcccc}
\toprule
\textbf{Model} & \textbf{Full} & \textbf{0.7$\times$} & \textbf{0.6$\times$} & \textbf{0.5$\times$} \\
\midrule
Balcony-Llama3-8B & 58.16 & 41.82 \scriptsize(1.4$\times$) & 35.26 \scriptsize(1.7$\times$) & 30.37 \scriptsize(1.9$\times$) \\
Balcony-Llama2-7B & 52.54 & 37.23 \scriptsize(1.4$\times$) & 32.46 \scriptsize(1.6$\times$) & 27.99 \scriptsize(1.9$\times$) \\
\bottomrule
\end{tabular}
}
\end{table}

\begin{table*}[htb]
\centering
\caption{Performance Comparison of pretraining from scratch and tuning a Pretrained model for Balcony and Sorted Approaches. For pretraining, both the Balcony and Sorted methods use 384B tokens. In the standard approach, 15B tokens are used to tune the baseline. In the Balcony method, the base model is frozen while in Sorted approach, the entire model is updated. The reported numbers represent 5-shot performance on MMLU, 10-shot performance on HellaSwag and zero-shot on other tasks. The notation \#Params refers to the number of non-embedding parameters.} 
%The numbers represent accuracy on ARC \citet{clark2018think}, BoolQ \citet{}, OpenbookQA \citet{}, PIQA \citet{Bisk2020}, and WinoGrande \citet{sakaguchi2021winogrande}, as well as 5-shot performance on MMLU \citet{hendrycks2020measuring} and 10-shot performance on HellaSwag \citet{zellers2019hellaswag}. The notation \#Params refers to the number of non-embedding parameters.}
\label{table:performance_pretraining}
\resizebox{\textwidth}{!}{
\begin{tabular}{l |l c c c c c c c c c c c c}
\toprule
& \textbf{Model} & \textbf{\# Params} & \textbf{ARC-C} & \textbf{ARC-E} & \textbf{BoolQ} & 
\textbf{Lambada-Op} & 
\textbf{Lambada-ST} & 
\textbf{OpenBookQA} &
\textbf{PIQA} & \textbf{Winogrande} & 
\textbf{HellaSwag(10)} & \textbf{MMLU(5)} &  \textbf{Avg} \\
\cmidrule{2-14}
&\textbf{LLM-1B (Baseline)}& 973M & 37.6 & 74.3 & 58.4 & 50.4 & 43.3 & 30 & 74.9 & 59.8 & 46.4 & 25.9 & 50.1  \\
\cmidrule{2-14}
%973M & 31.3 & 65.5 & 64 & 63 & 54 & 26.4 & 74.5 & 60.69 & 48.4 & 32.1 & 52 \\
\multirow{8}{*}{\rotatebox{90}{\textbf{Pretraining}}} 
&\textbf{Balcony-XL}  & 973M & 38.1 & 72.7 & 60 & 47.9 & 40.2 & 27.8 & 74.4 & 56.2 & 45.5 & 25.1 & \textbf{48.8} \\
&\textbf{Sorted-XL} & 973M & 36.6 & 71.8 & 59.8 & 43.6 & 35.6 & 28.8 & 72.8 & 53.67 & 43.5 & 25.6 & 47.2 \\
\cmidrule{2-14}
&\textbf{Balcony-L}  & 790M & 37.8 & 72 & 61.8 & 47.4 & 40 & 27.2 & 74.4 & 55.17 & 44.4 & 25.3 & \textbf{48.6 }\\
&\textbf{Sorted-L} & 729M & 35.3 & 71.8 & 60.9 & 43.1 & 35.6 & 28.6 & 71.8 & 54.38 & 42.5 & 24.8 & 46.9 \\
\cmidrule{2-14}
&\textbf{Balcony-M} & 547M & 33.4 & 69.1 & 62.9 & 46.2 & 37.8 & 26.4 & 71.5 & 56.75 & 41.7 & 26.1 & \textbf{47.2} \\
&\textbf{Sorted-M}  & 486M & 32.1 & 68.1 & 60.3 & 42.2 & 32.7 & 26.8 & 70.4 & 54.06 & 40.4 & 24.8 & 45.2 \\
\cmidrule{2-14}
&\textbf{Balcony-S}  & 304M & 27 & 63.6 & 60.5 & 36.3 & 24.2 & 23.2 & 70.5 & 50.7 & 35.7 & 25.9 & \textbf{41.8 }\\
&\textbf{Sorted-S}  & 243M & 26.5 & 61.6 & 60 & 31.8 & 20 & 21 & 67.4 & 50.59 & 34 & 27.1 & 40 \\
\midrule
\midrule

\multirow{8}{*}{\rotatebox{90}{\textbf{Standard}}} 
&\textbf{Balcony-XL} & 973M & 37.6 & 74.3 & 58.4 & 50.4 & 43.3 & 30 & 74.9 & 59.8 & 46.4 & 25.9 & \textbf{50.1 } \\ 
&\textbf{Sorted-XL} & 973M & 34.4 &	68.1 & 49.9 & 46.3	& 34.8	& 27 & 	72.2 & 56.7 & 43.2 & 25.7 & 45.8  \\
\cmidrule{2-14}
&\textbf{Balcony-L}& 790M & 32.8 & 66.3 & 61.4 & 45.1 & 36.4 & 25.4 & 71.3 & 57.6 & 41.3 & 26.3 & \textbf{46.4 }\\
&\textbf{Sorted-L} & 729M &  30.5 &57.5	& 53.4	& 39.2	& 26.7	& 22.4	& 67.4	& 53.7	& 36.8	& 25.5	& 41.3	\\ 
\cmidrule{2-14}
&\textbf{Balcony-M}& 547M  & 24.9 & 58.7 & 61.4 & 29.5 & 20.5 & 21.8 & 67.8 & 53.4 & 34.3 & 27.2 & \textbf{40} \\ 
&\textbf{Sorted-M} & 486M & 22.9& 44.2	& 43.4	& 20.2	& 13	& 17.8 & 	63.7	& 52.6 & 	30.4	& 25.5 &	33.4 \\ 
\cmidrule{2-14}
&\textbf{Balcony-M}& 304M  & 22.3 & 51.7 & 62  & 19.3 & 8.2  & 17.4 & 64.1 & 52.2 & 30.3 & 24.6 & \textbf{35.2 } \\
&\textbf{Sorted-S} & 243M  & 21.2	& 40.8	& 55.6	& 11.3 & 4.4 & 	13.6 & 60.3	& 49.1	& 27.9	& 25.6	& 31  \\ 

\bottomrule

\end{tabular}
}
\end{table*}

\vspace{-2pt}
\subsection{Ablation studies}
\label{sec:ablation}

\smallskip\noindent \textbf{Pretraining from scratch} In this experiment, we train both the base model and the Balcony layers from scratch to assess the representational capacity of Balcony compared to a nested design. This approach is particularly useful when the entire pretraining budget is allocated to developing a dynamic model.

We begin by training the LLM-1B model from scratch, which serves as the baseline for normal training (see the Training Details in \ref{sec:experiments_setup}). Next, we train the Balcony layers alongside the base model using the same training budget. The Balcony layers are initialized randomly, and since we are training from scratch, we omit self-distillation. Instead, we use the average loss over all submodels as the training objective. The Balcony exit layers are placed after layers 4, 8, 12, and 16 of the model.

We also compare Balcony’s representational capacity to SortedLLaMA, which applies sorting training only at the fine-tuning stage. To ensure a fair comparison, we perform the same pretraining with SortedLLaMA, but with a sorted objective function. The results, shown in Table \ref{table:performance_pretraining}, indicate that for all submodels, pretraining using the Balcony design outperforms the nested approach used in SortedLLaMA.  
However, it is important to note that when training from scratch, since the base model is not frozen, the accuracy of the resulting model, despite outperforming SortedLLaMA, is lower than the baseline model.

Furthermore, we compare pretraining with the standard efficient training method proposed by Balcony and evaluate it against tuning using the Sorted approach. Note that in Balcony, the base model is frozen, and only the Balcony module is updated, whereas in the Sorted approach, the entire model is updated. The results show that Balcony provides significantly higher accuracy across submodels compared to the Sorted approach.

\begin{figure*}[h!]
  \centering
  \hspace{-10pt}
  \begin{minipage}[b]{0.34\textwidth}
    \centering
    \includegraphics[width=\columnwidth]{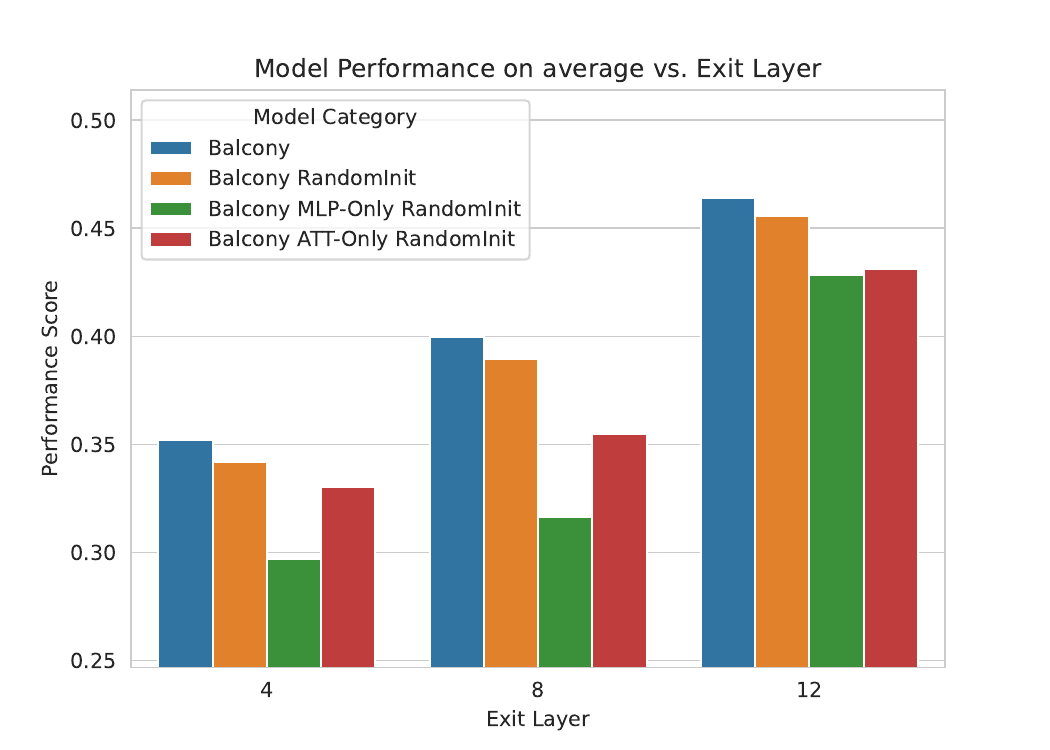}
  \end{minipage} 
  \hspace{-10pt}
  \begin{minipage}[b]{0.34\textwidth}
    \centering
    \includegraphics[width=\columnwidth]{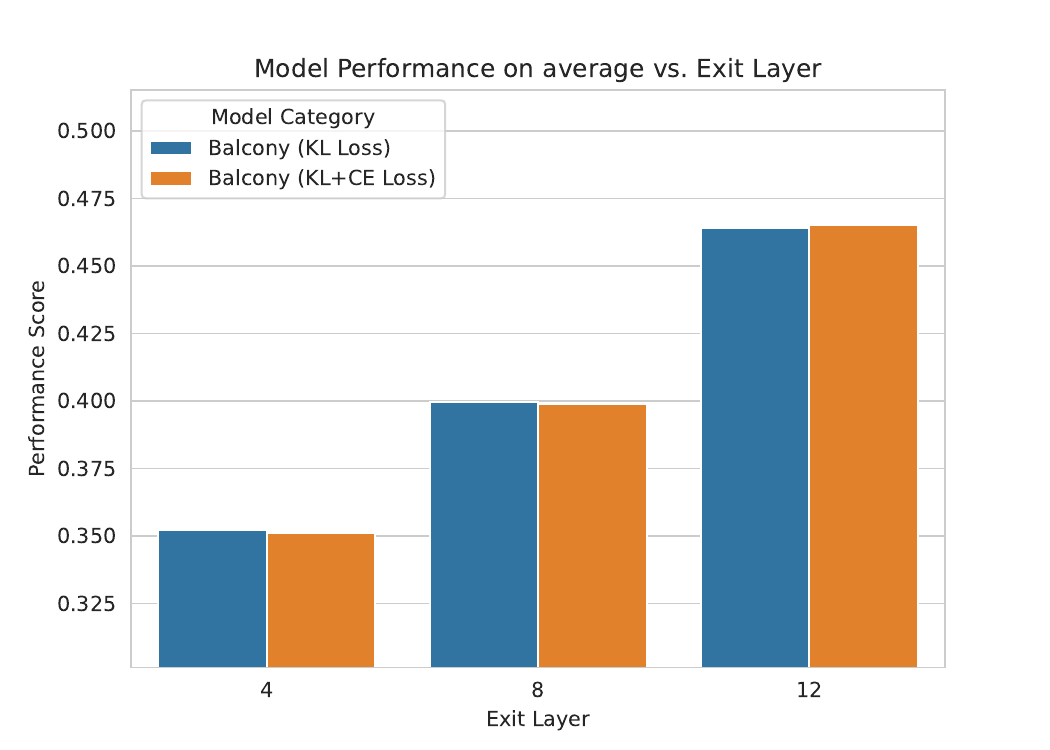}
  \end{minipage} 
  \hspace{-10pt}
  \begin{minipage}[b]{0.34\textwidth}
    \centering
    \includegraphics[width=\columnwidth]{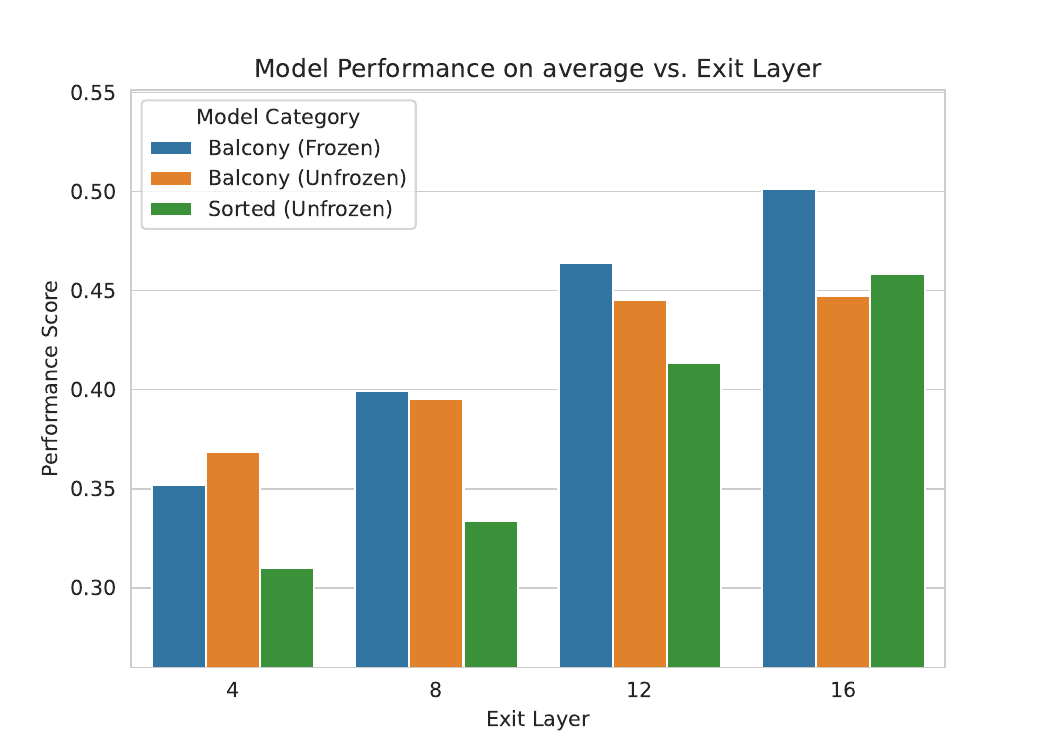}
  \end{minipage}

  \caption{Ablation studies on the Balcony framework. From left to right: (1) Impact of random initialization and the contributions of MLP-only and Attention-only modules. (2) Effect of incorporating Cross-Entropy (CE) loss in self-distillation. (3) Effect of freezing the main architecture during post-training. The results are the average scores across ARC-E, ARC-C, BoolQ, HellaSwag, Lambada, MMLU, OpenBookQA, PIQA, and Winogrande.}
  \label{fig:ablations}
\end{figure*}
\vspace{-2pt}

\smallskip\noindent \textbf{Random Initialization, MLP-Only and Attention-Only Balcony}
A key question that arises is why we use a single transformer decoder layer and why we initialize it from the final layer of the full architecture. To investigate this, we conduct an ablation study by repeating the Balcony-LLM-1B experiment under different configurations to assess the impact of each module. First, we perform balcony training with randomly initialized balcony modules (transformer decoders) to evaluate the effect of initializing from the final layer on submodel performance. Additionally, we train balcony models with MLP-only and Attention-only modules, using the same set of intermediate layers in the LLM-1B model, to isolate the contributions of each component. Figure \ref{fig:ablations} presents the average results of these submodels across ARC-E, ARC-C, BoolQ, HellaSwag, Lambada, MMLU, OpenBookQA, PIQA, and Winogrande benchmarks. As shown, standard balcony training with a transformer decoder initialized from the final layer outperforms both random initialization and the MLP-only and Attention-only variants.

% \begin{figure}[t]
%   \centering
%   \includegraphics[width=\columnwidth]{ablations/ce_kl/average.pdf}
%   \caption{Ablation study on the impact of including CE loss in self-distillation. The results are the average score across ARC-E, ARC-C, BoolQ, HellaSwag, Lambada, MMLU, OpenBookQA, PIQA, and Winogrande benchmarks.}
%   \label{fig:ce_kl}
% \end{figure}
\smallskip\noindent \textbf{Effect of Cross-Entropy Loss}
During post-training of the balcony modules, we used only KL-divergence loss between the frozen, pretrained full model’s output and the outputs of the balcony modules. To examine the potential effect of incorporating Cross-Entropy (CE) loss in training balcony modules, we conducted an ablation study by training them with both KL-divergence and CE loss on the submodels of our LLM-1B model. In this experiment, we set the KL loss weight to 0.001. Figure \ref{fig:ablations} presents the results of standard balcony training (KL-only) and the KL + CE variant across different submodels and benchmarks. As shown, incorporating CE loss does not yield significant improvements in the performance of balcony submodels.

% \begin{figure}[t]
%   \centering
%   \includegraphics[width=\columnwidth]{ablations/freezing/average.pdf}
%   \caption{Ablation study on the impact of freezing the main architecture during post training. The results are the average score across ARC-E, ARC-C, BoolQ, HellaSwag, Lambada, MMLU, OpenBookQA, PIQA, and Winogrande benchmarks.}
%   \label{fig:freezing}
% \end{figure}
\smallskip\noindent \textbf{Effect of Freezing during Balcony Training}
During balcony training, we kept the pretrained backbone model weights frozen. To assess the impact of updating the main model’s weights alongside the balcony modules, we conducted an ablation study where the backbone model's weights were also made trainable. Additionally, we included another dynamic inference baseline that does not freeze the backbone model’s weights: Sorted Fine-Tuning \cite{kavehzadeh2024sorted}. Figure \ref{fig:ablations} presents the results of three setups: standard balcony training (frozen backbone), balcony training with an unfrozen backbone, and Sorted Fine-Tuning (unfrozen backbone) on different submodels of the LLM-1B model. As observed, training balcony modules while keeping the backbone model frozen not only preserves the performance of the full pretrained architecture but also outperforms both the unfrozen balcony and Sorted Fine-Tuning approaches across most submodels. More detailed results can be found in the appendix.%A more comprehensive set of ablation studies covering all models and different training strategies is provided in the appendix. %These additional experiments further validate our findings by analyzing the impact of various design choices and training configurations.

\vspace{-3pt}
\section{Conclusion}
\vspace{-3pt}
In this paper, we presented Balcony, a novel framework for depth-based dynamic inference that addresses key challenges in the deployment of large-scale language models under strict computational constraints. By introducing additional transformer layers at selected exit points and training them with a self-distillation loss, Balcony enables flexible, real-time adaptation to various computational budgets without compromising the performance of the full model. Our experiments demonstrate that Balcony achieves minimal performance degradation while using a small training budget. This efficiency represents a significant improvement over state-of-the-art methods. Balcony’s simplicity, effectiveness, and minimal resource requirements position it as a promising approach to dynamic inference, paving the way for more efficient deployment of large-scale models across diverse hardware environments. Future work will explore extending Balcony and combining it with more efficient architectures, such as MoE and state space models like Mamba, to further enhance performance and scalability. Additionally, Balcony can be leveraged for self speculative decoding \citep{leviathan2023fast, chen2023accelerating} to achieve speedup without degradation of model performance. 

\section{Limitations}
%Some of the most powerful emerging models are notably deep \citet{chen2021evaluatinglargelanguagemodels, deepseekai2024deepseekllmscalingopensource,bai2023qwen,bi2024deepseek}, and the potential of the proposed framework increases with model depth. However, due to computational budget constraints, extensive experiments on deeper models could not be conducted.

While Balcony minimizes performance degradation at reduced latencies, there remains an inherent trade-off between latency and model accuracy. For extremely low-latency requirements, further performance degradation might be inevitable and combining Balcony with width compression methods may become necessary. 

While Balcony can be easily combined with certain compression strategies, like quantization, we have not explored how it interacts with techniques (e.g., low-rank factorization or pruning). Understanding these interactions could further optimize memory and speed.

Also, in our experiments, we only evaluate the model at predefined budgets. However, implementing token-level confidence-based exits could potentially yield better trade-offs between accuracy and latency, and help boost our performance even further.

%\section*{Acknowledgments}
\bibliography{balcony}

\newpage 

\appendix
\section{Benchmark Descriptions}

This section provides a brief overview of the benchmarks used for evaluation.

\subsection{ARC-E (AI2 Reasoning Challenge - Easy)}
ARC-E consists of multiple-choice science questions designed to evaluate a model's reasoning ability \cite{clark2018think}. The dataset contains 7,787 questions, with 2,251 categorized as 'easy.' These questions require commonsense reasoning beyond simple retrieval.

\subsection{LAMBADA}
LAMBADA evaluates a model’s ability to predict the last word of a passage, requiring deep contextual understanding \cite{paperno2016lambada}. It consists of 10,022 book passages, where the final word is only guessable with full comprehension.

\subsection{PIQA (Physical Commonsense Reasoning)}
PIQA assesses a model’s ability to reason about physical interactions \cite{bisk2020piqa}. The dataset contains 16,000 training and 2,000 validation questions, evaluating how well models understand everyday physical tasks.

\subsection{Winogrande}
Winogrande is a large-scale dataset designed for commonsense reasoning using Winograd-style problems \cite{sakaguchi2020winogrande}. It contains over 44,000 sentence-pair problems that require resolving ambiguous pronouns using contextual knowledge.

\subsection{MMLU 5 (Massive Multitask Language Understanding)}
MMLU tests knowledge and reasoning across 57 domains \cite{hendrycks2021measuring}. The MMLU-5 subset focuses on five key categories to assess broad cognitive abilities through multiple-choice questions ranging from elementary to expert level.

\subsection{HellaSwag}
HellaSwag \citet{hellaswag} is a large-scale  reasoning benchmark that tests the ability of models to predict the most likely continuation of a sentence in everyday situations. The dataset contains multiple-choice questions designed to challenge models on contextual reasoning and world knowledge. 

\section{Ablations}

In order to select the best architecture for the \textbf{Balcony} model, ablation studies were conducted on different loss functions, activation functions, and parameter initialization methods.

\subsection{Effect of CE and KL Losses}

In the Balcony architecture, two possible loss functions can help align the Balcony layer’s output with that of the full model. One end-to-end approach is to match the generated output using the \textbf{Cross-Entropy (CE)} loss. Another approach is to apply a distillation objective by aligning the logits of the full model and the Balcony layer using the \textbf{Kullback-Leibler (KL)} divergence loss.

\begin{figure*}[h]
\caption{Ablation study on the impact of including CE loss in self-distillation. The results are the score across ARC-E, ARC-C, BoolQ, HellaSwag, Lambada, MMLU, OpenBookQA, PIQA, and Winogrande benchmarks.}
  \label{fig:ce_kl}
    \centering
    \begin{minipage}{0.32\textwidth}
        \includegraphics[width=\linewidth]{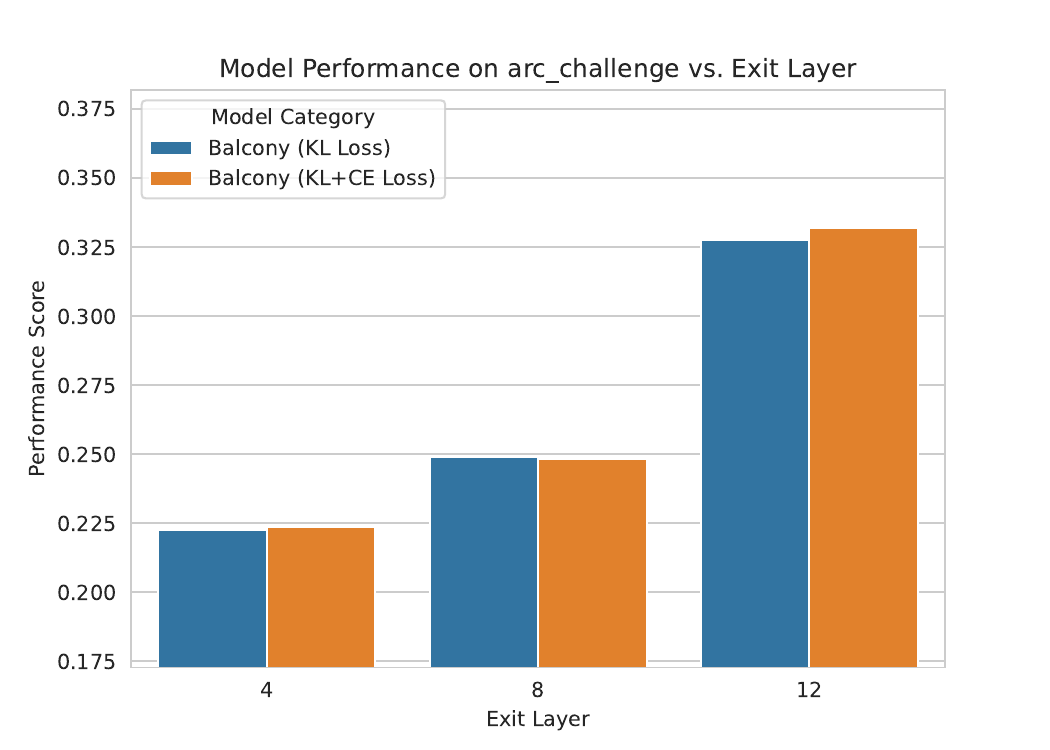}
    \end{minipage}
    \begin{minipage}{0.32\textwidth}
        \includegraphics[width=\linewidth]{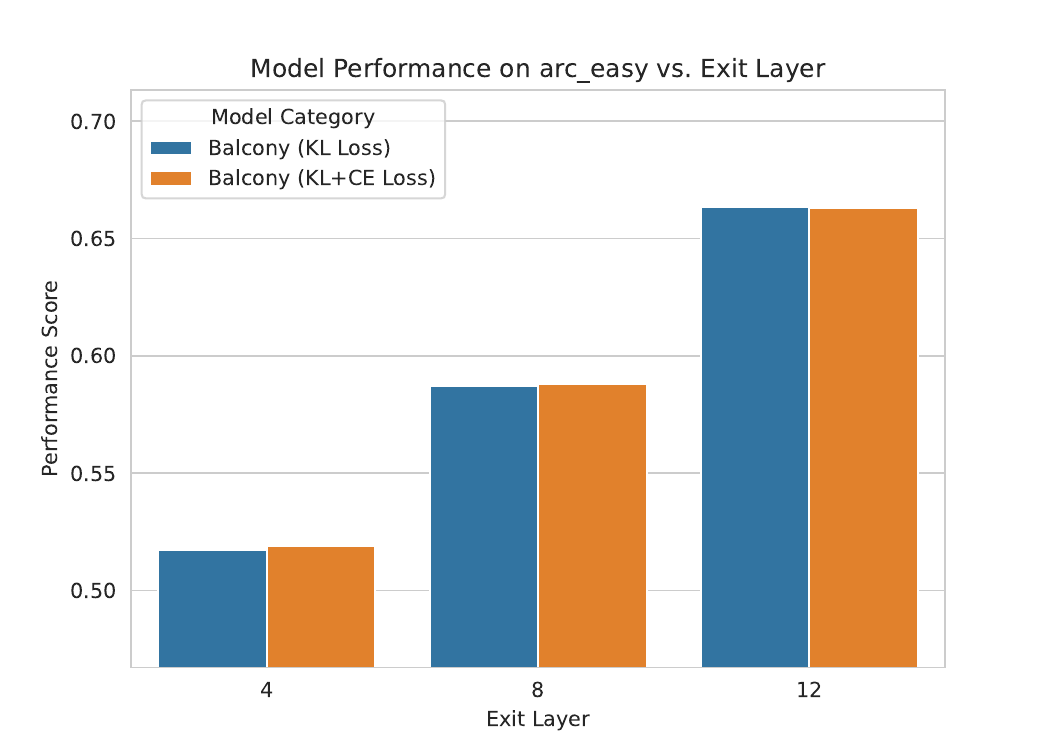}
    \end{minipage}
    \begin{minipage}{0.32\textwidth}
        \includegraphics[width=\linewidth]{ablations/ce_kl/average.pdf}
    \end{minipage}
    \begin{minipage}{0.32\textwidth}
        \includegraphics[width=\linewidth]{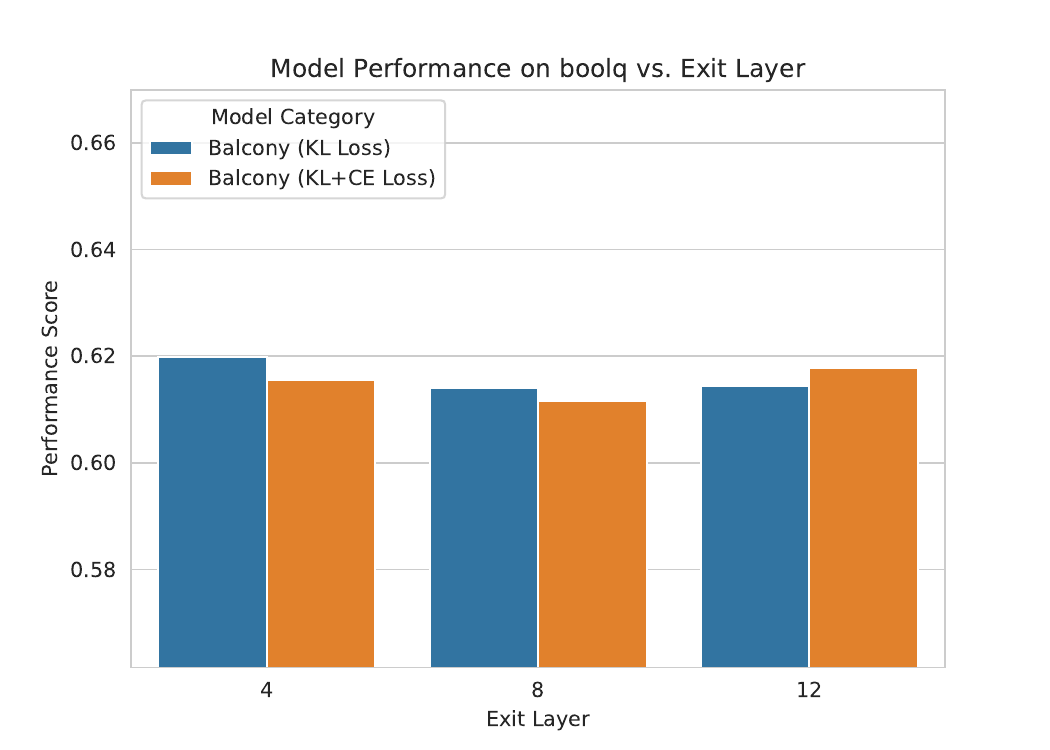}
    \end{minipage}
    \begin{minipage}{0.32\textwidth}
        \includegraphics[width=\linewidth]{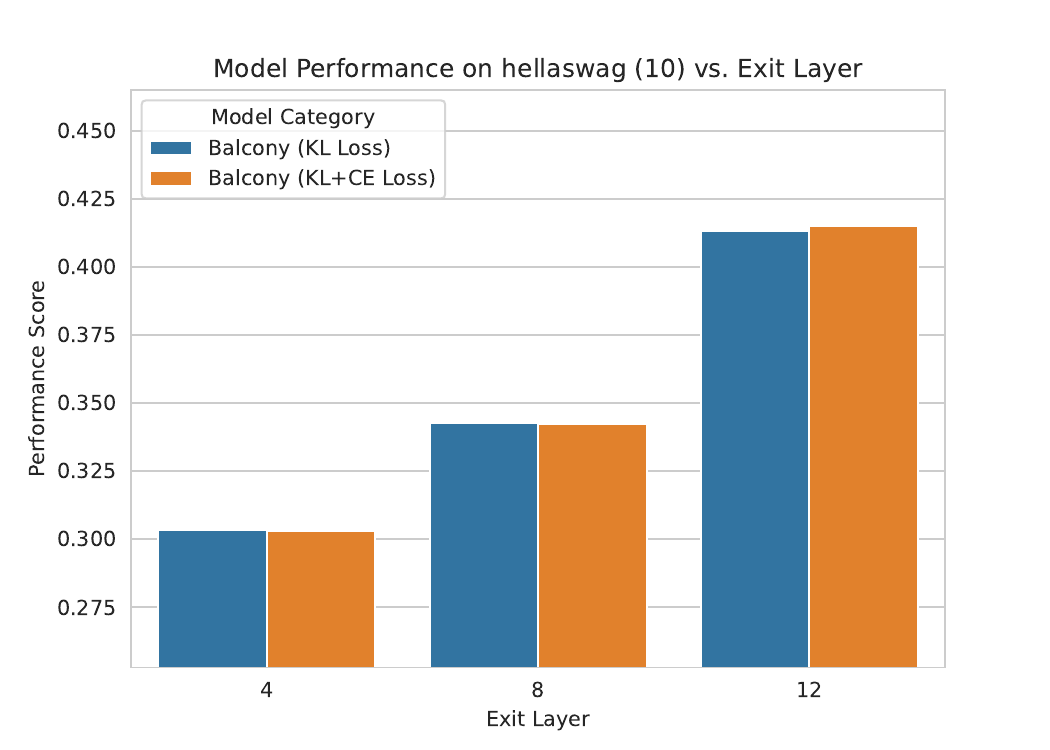}
    \end{minipage}
    \begin{minipage}{0.32\textwidth}
        \includegraphics[width=\linewidth]{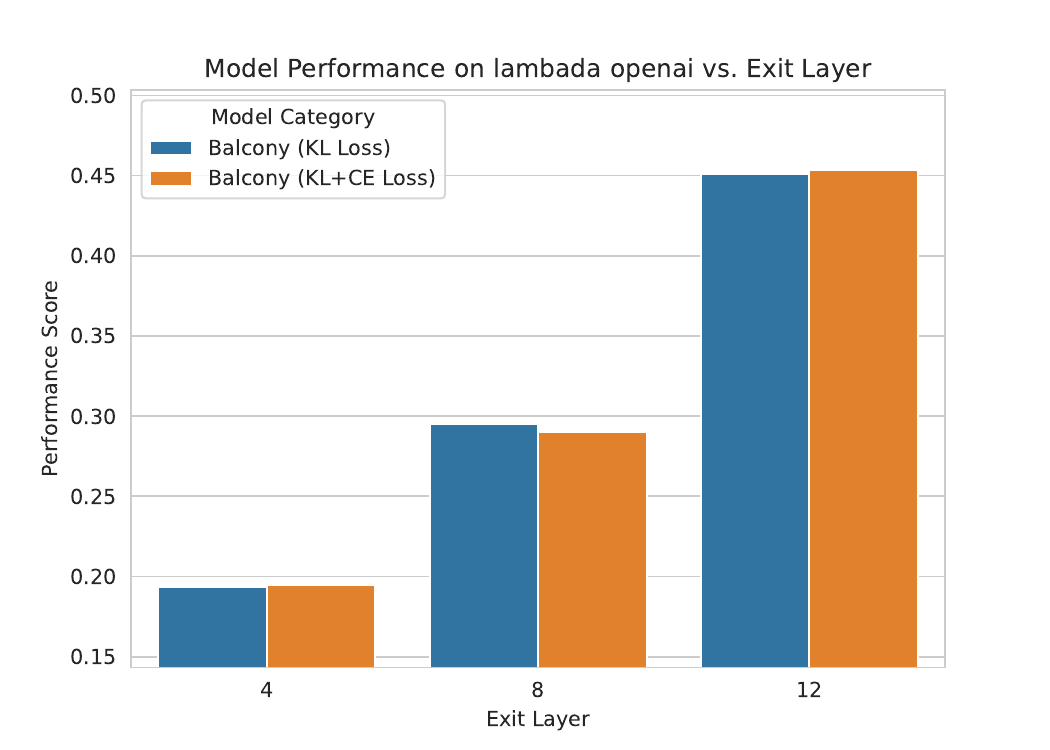}
    \end{minipage}
    \begin{minipage}{0.32\textwidth}
        \includegraphics[width=\linewidth]{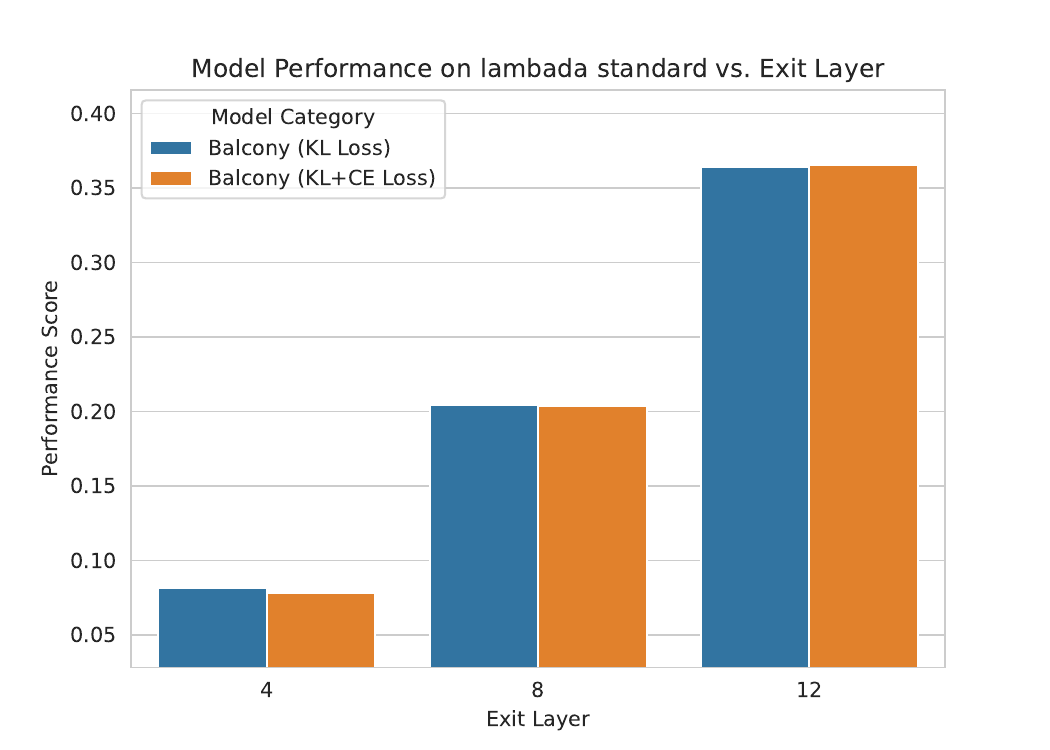}
    \end{minipage}
    \begin{minipage}{0.32\textwidth}
        \includegraphics[width=\linewidth]{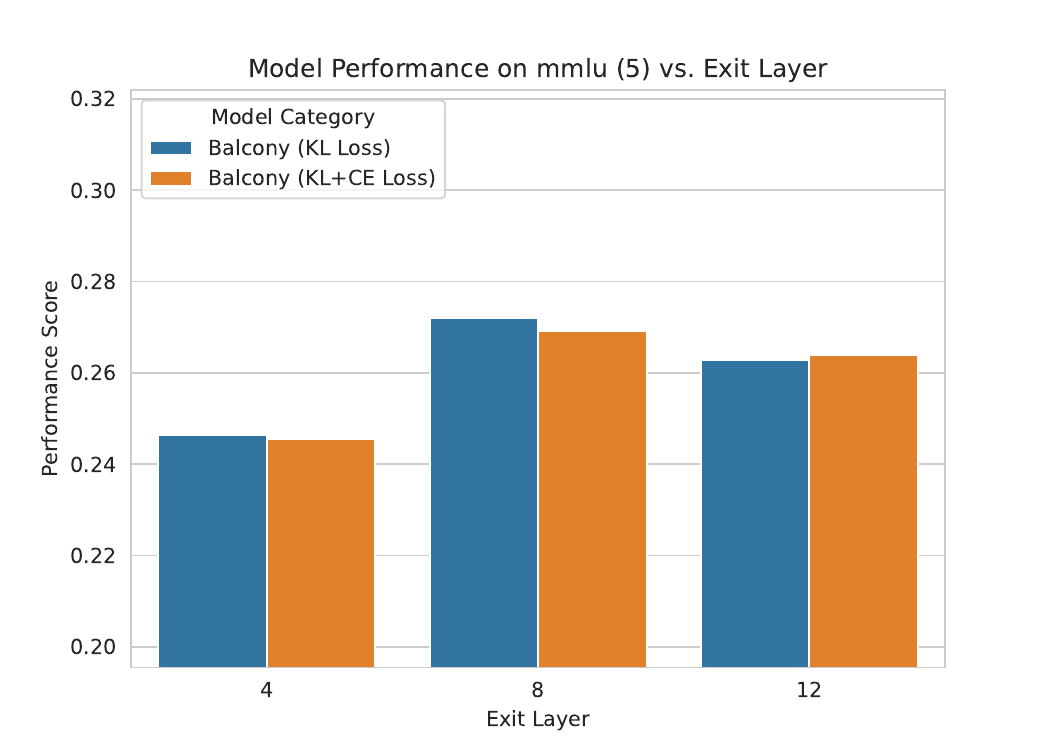}
    \end{minipage}
    \begin{minipage}{0.32\textwidth}
        \includegraphics[width=\linewidth]{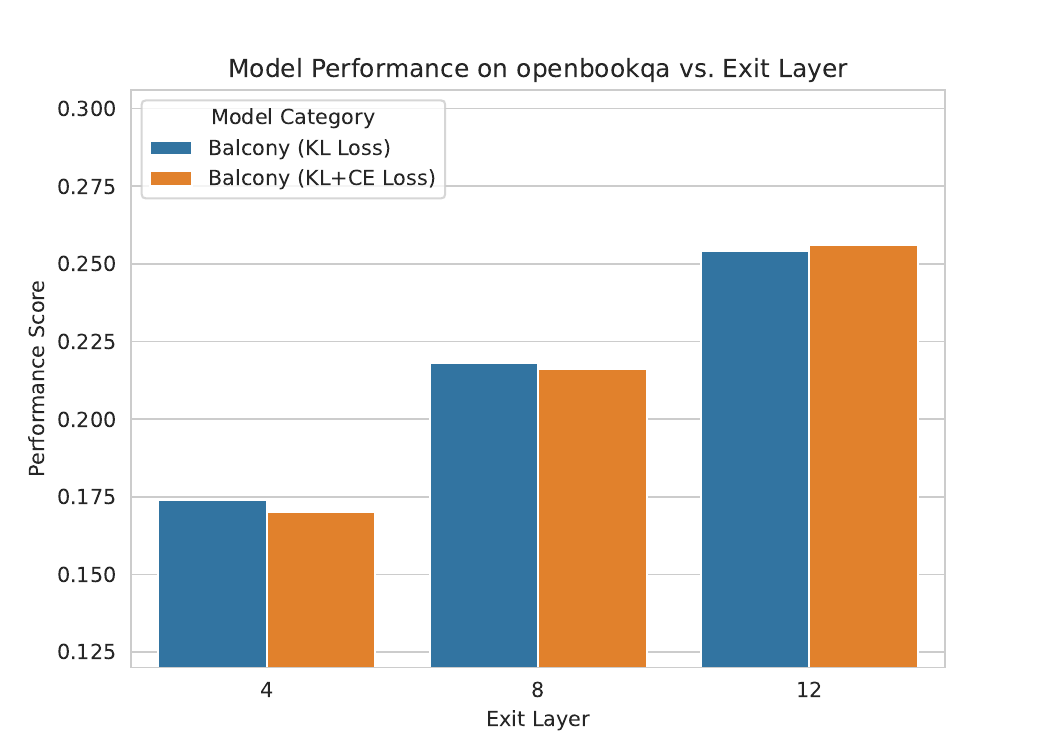}
    \end{minipage}
    \begin{minipage}{0.32\textwidth}
        \includegraphics[width=\linewidth]{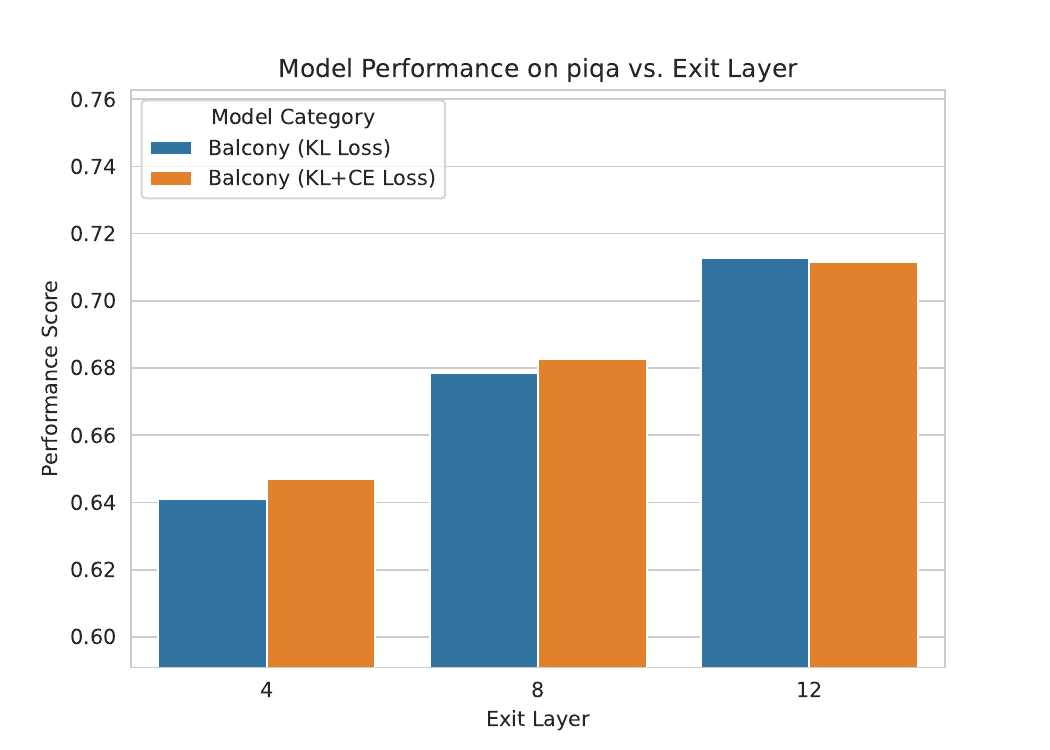}
    \end{minipage}
    \begin{minipage}{0.32\textwidth}
        \includegraphics[width=\linewidth]{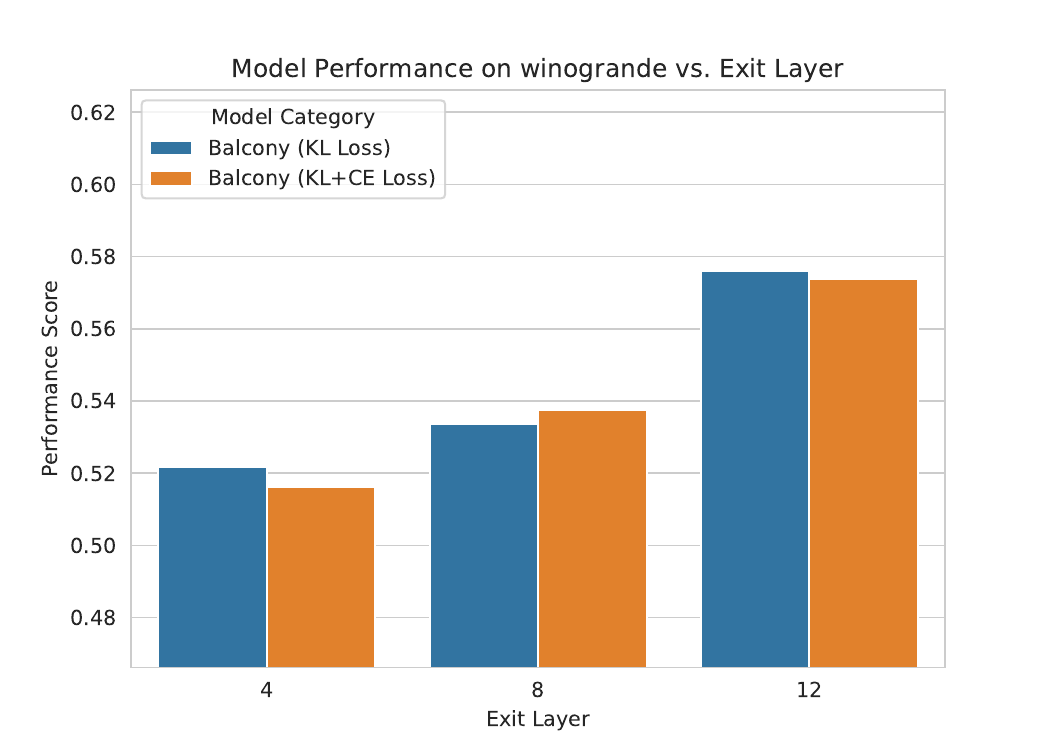}
    \end{minipage}
\end{figure*}

From Figure~\ref{fig:ce_kl}, we observe that combining both KL divergence and CE loss leads to a slight but consistent decrease in performance across different exit layers and benchmarks.

\subsection{Effect of Freezing}
Depending on the training budget, different layers of the architecture can be either frozen or unfrozen. To analyze the trade-off between performance and training cost, we report benchmark evaluation scores for both the frozen and unfrozen versions of the main model alongside the Balcony layer. Additionally, we compare these results with Sorted~\cite{kavehzadeh2024sorted}, which shares the same base structure as Balcony but without exit layers.

\begin{figure*}[h]

  \caption{Ablation study on the impact of freezing the main architecture during post training. The results are the score across ARC-E, ARC-C, BoolQ, HellaSwag, Lambada, MMLU, OpenBookQA, PIQA, and Winogrande benchmarks.}
  \label{fig:freezing}
    \centering
    \begin{minipage}{0.32\textwidth}
        \includegraphics[width=\linewidth]{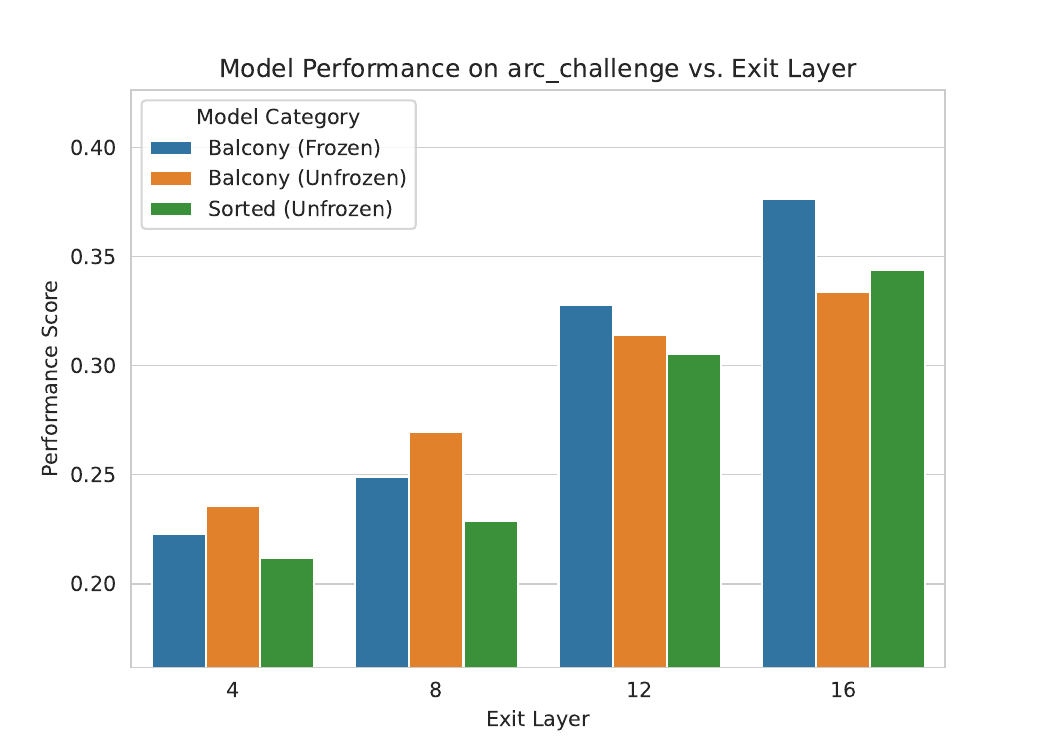}
    \end{minipage}
    \begin{minipage}{0.32\textwidth}
        \includegraphics[width=\linewidth]{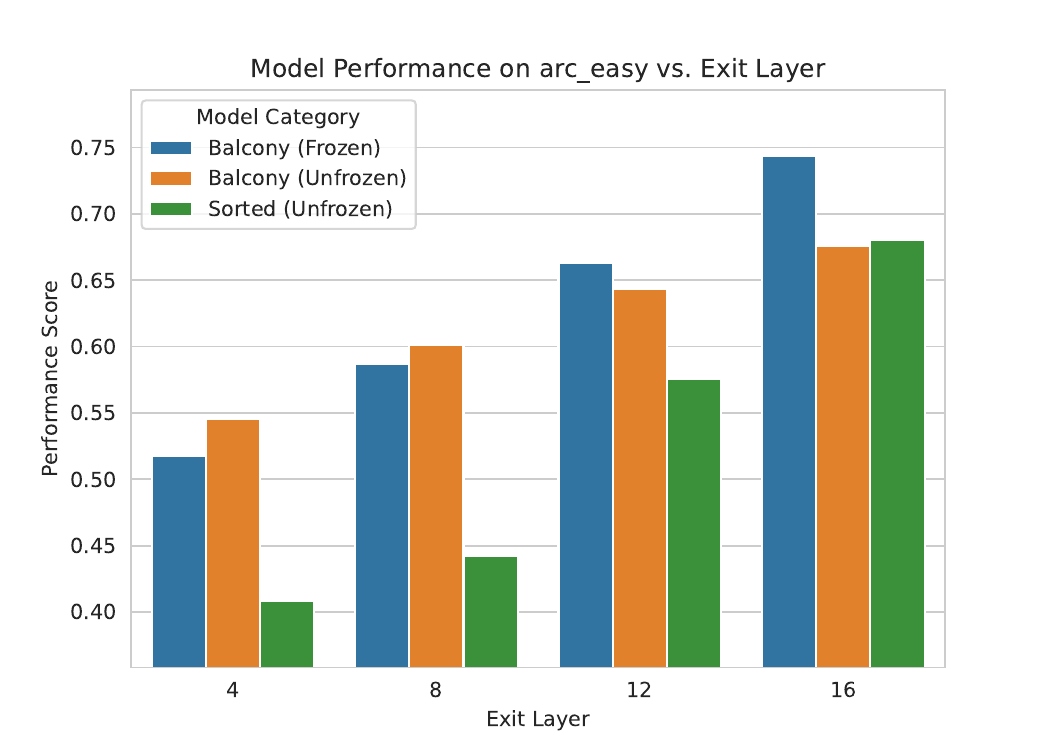}
    \end{minipage}
    \begin{minipage}{0.32\textwidth}
        \includegraphics[width=\linewidth]{ablations/freezing/average.pdf}
    \end{minipage}
    \begin{minipage}{0.32\textwidth}
        \includegraphics[width=\linewidth]{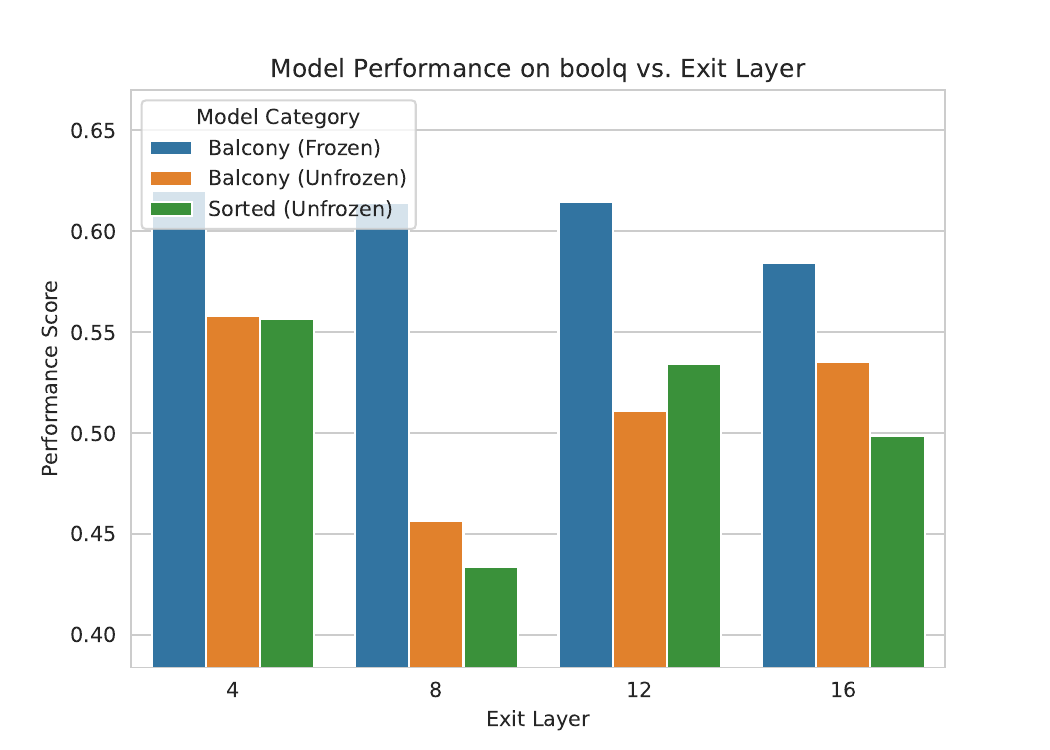}
    \end{minipage}
    \begin{minipage}{0.32\textwidth}
        \includegraphics[width=\linewidth]{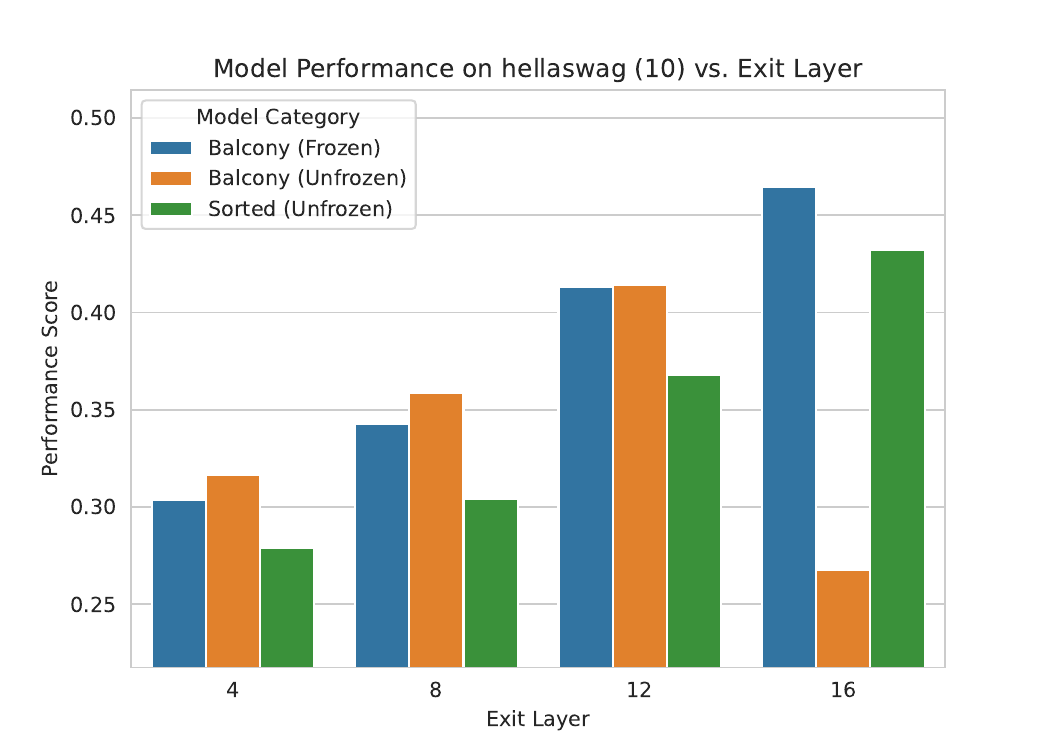}
    \end{minipage}
    \begin{minipage}{0.32\textwidth}
        \includegraphics[width=\linewidth]{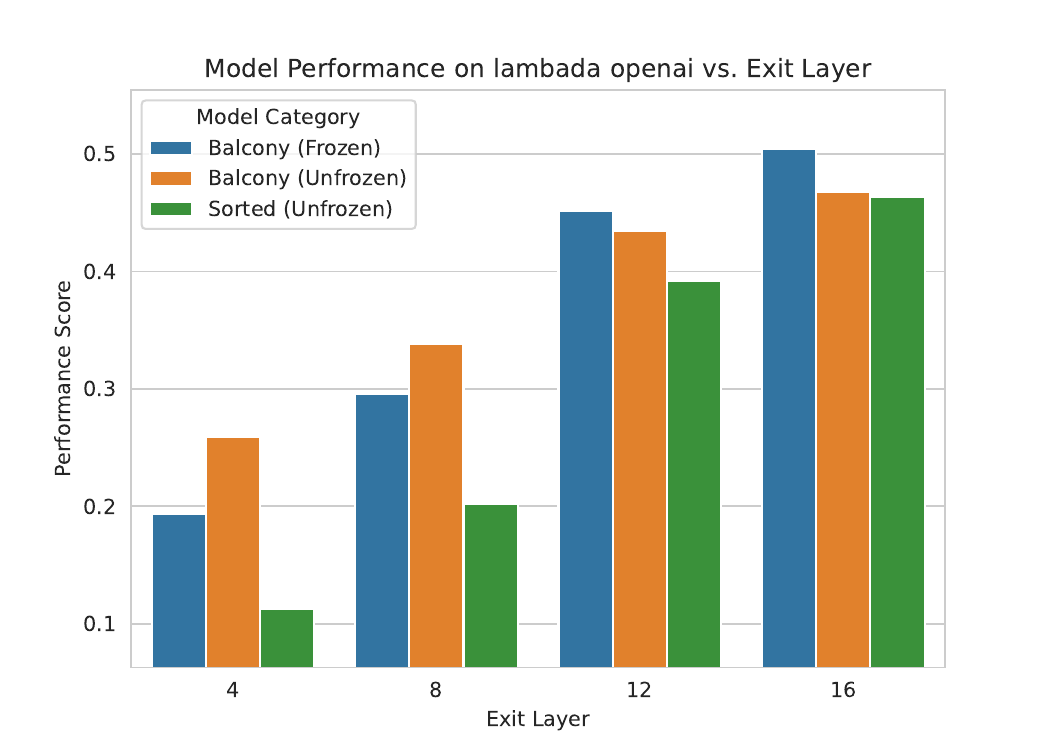}
    \end{minipage}
    \begin{minipage}{0.32\textwidth}
        \includegraphics[width=\linewidth]{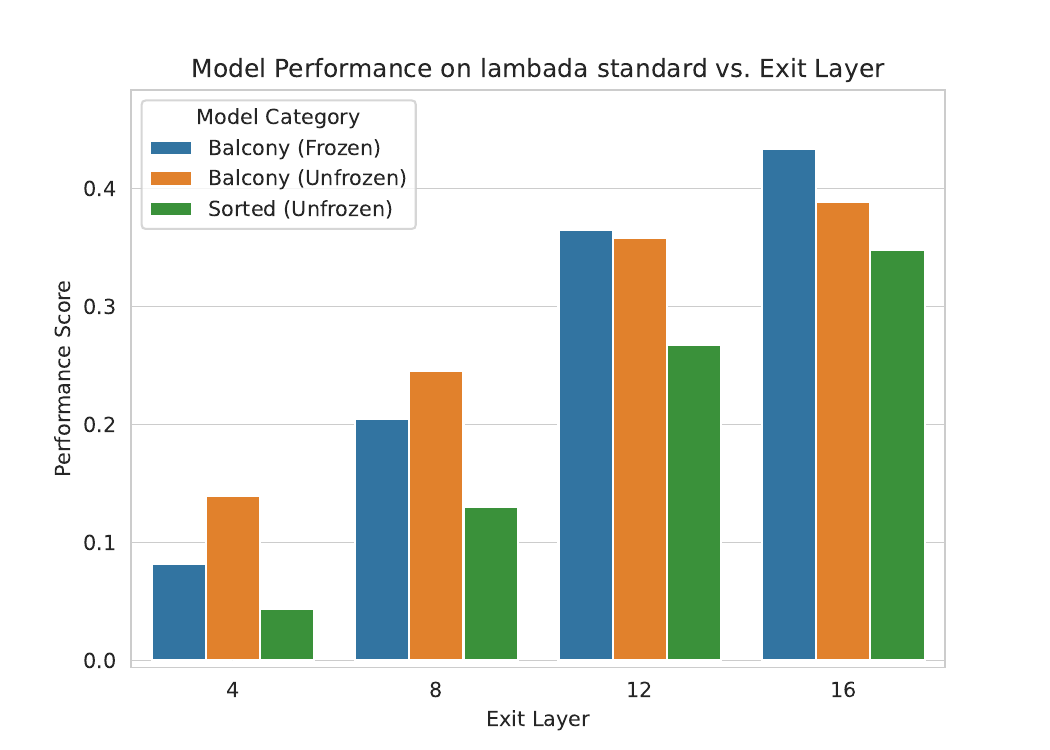}
    \end{minipage}
    \begin{minipage}{0.32\textwidth}
        \includegraphics[width=\linewidth]{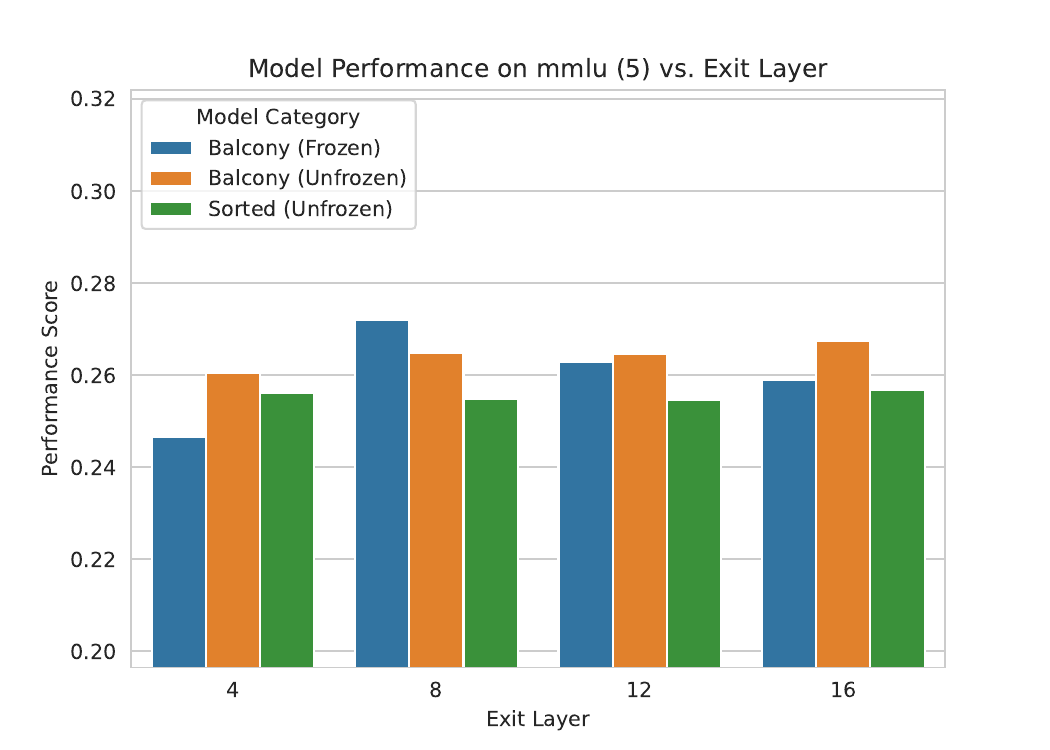}
    \end{minipage}
    \begin{minipage}{0.32\textwidth}
        \includegraphics[width=\linewidth]{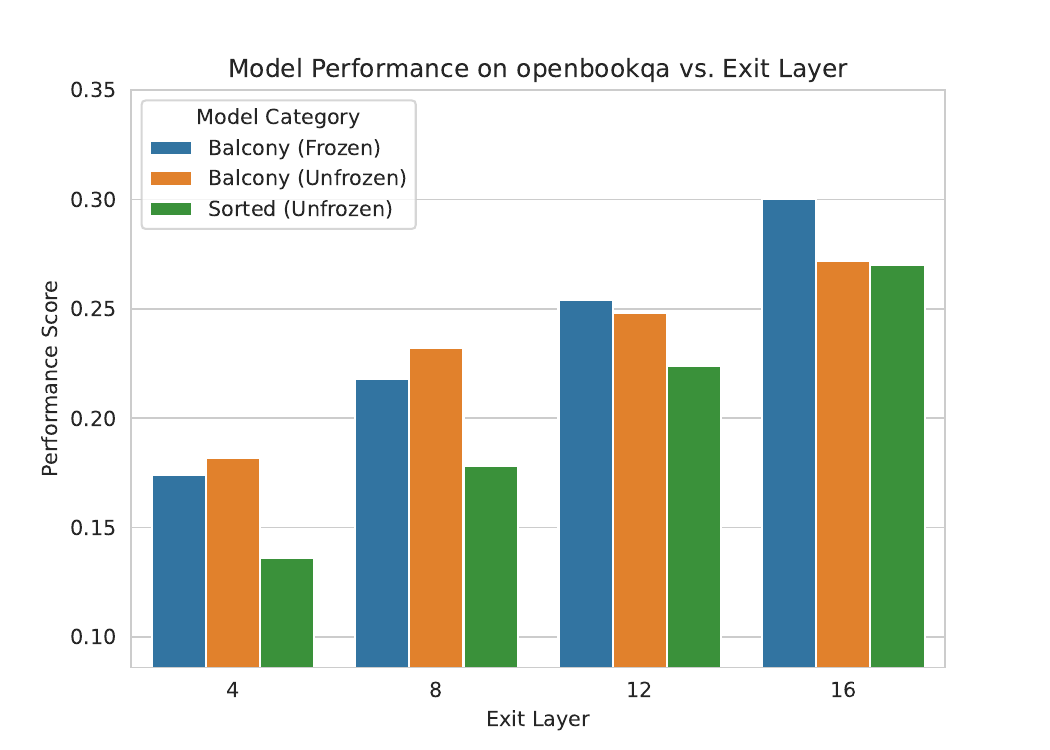}
    \end{minipage}
    \begin{minipage}{0.32\textwidth}
        \includegraphics[width=\linewidth]{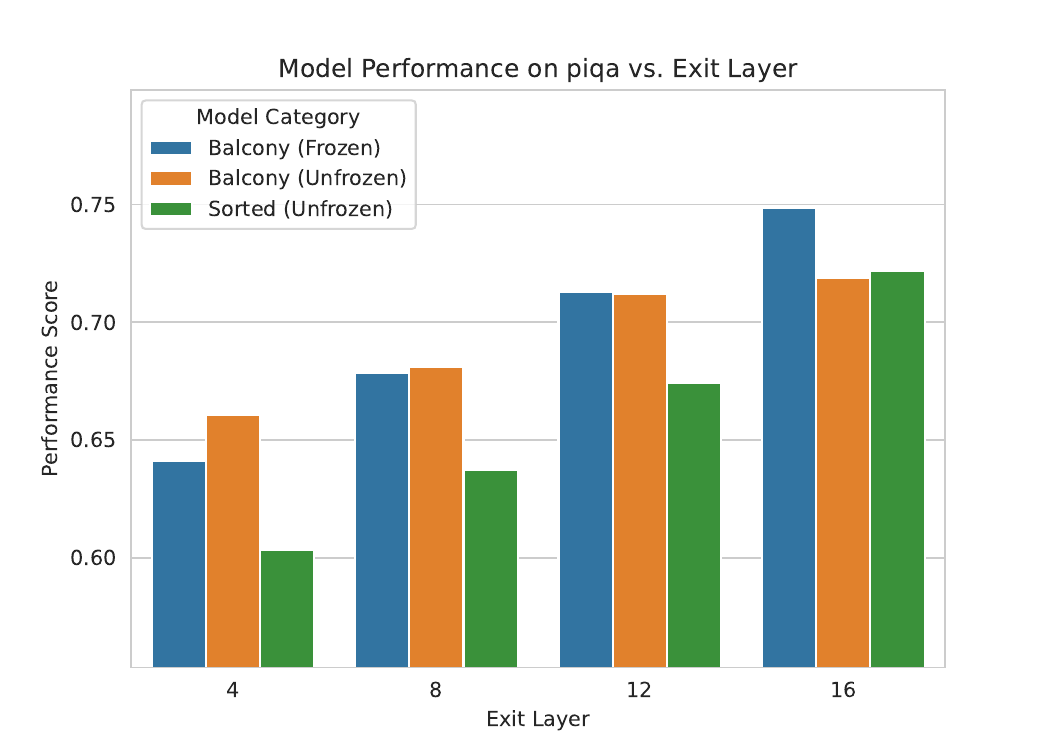}
    \end{minipage}
    \begin{minipage}{0.32\textwidth}
        \includegraphics[width=\linewidth]{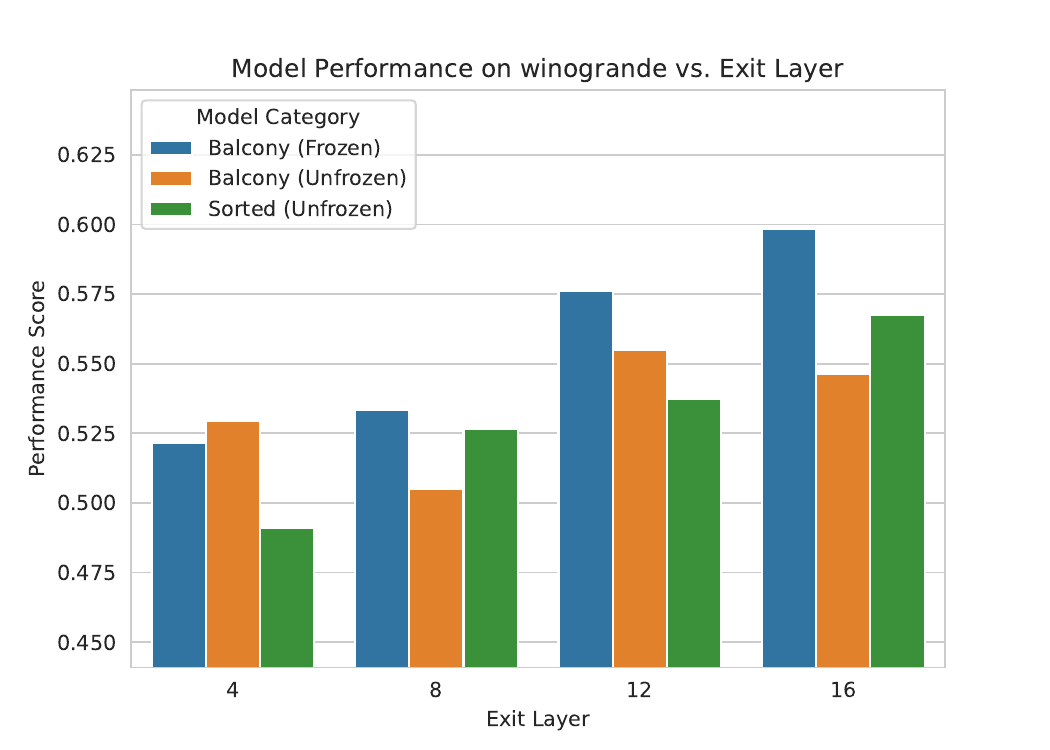}
    \end{minipage}
\end{figure*}

Interestingly, in Figure~\ref{fig:freezing} we observe that the frozen Balcony model outperforms its unfrozen counterparts. This result suggests that freezing layers may help mitigate issues such as catastrophic forgetting and overfitting, which often occur when fine-tuning a larger network.

\subsection{Effect of Random Inititialization}
The initialization of a generative model plays a crucial role in its performance, especially under a limited training budget. The Balcony architecture consists of multiple submodules, including MLP layers, attention layers, and the Balcony exit layer.

\begin{figure*}[h]
    \centering
    \caption{Ablation study on the impact of random initialization of balcony modules compared to regular balcony training starting from the full model final transformer layer weights. Also study on the effect of each MLP and Self-Attention modules in the balcony submodels' performance. The results are the score across ARC-E, ARC-C, BoolQ, HellaSwag, Lambada, MMLU, OpenBookQA, PIQA, and Winogrande benchmarks.}
    \label{fig:mlpattn}
    \begin{minipage}{0.32\textwidth}
        \includegraphics[width=\linewidth]{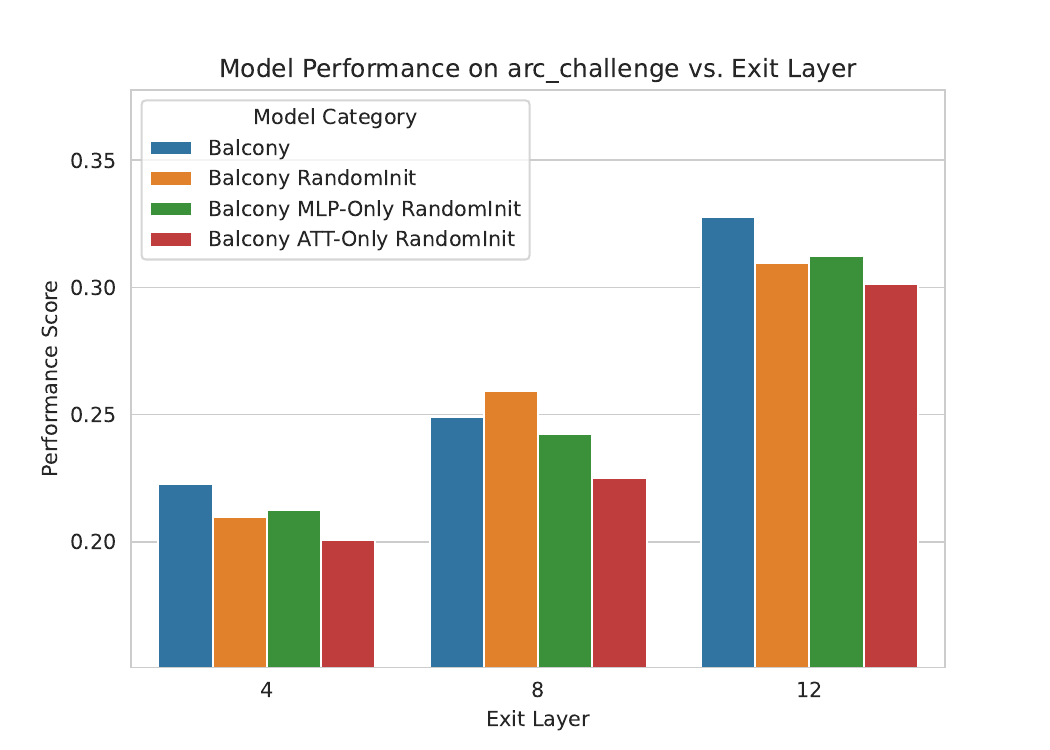}
    \end{minipage}
    \begin{minipage}{0.32\textwidth}
        \includegraphics[width=\linewidth]{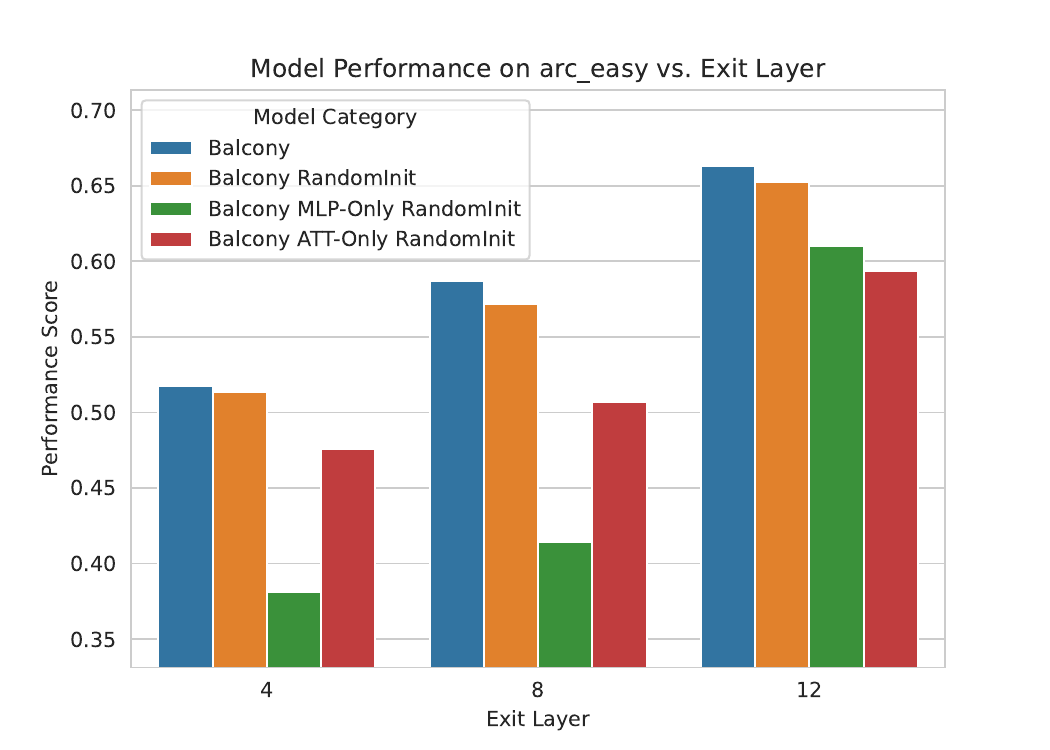}
    \end{minipage}
    \begin{minipage}{0.32\textwidth}
        \includegraphics[width=\linewidth]{ablations/randomInit/average.pdf}
    \end{minipage}
    \begin{minipage}{0.32\textwidth}
        \includegraphics[width=\linewidth]{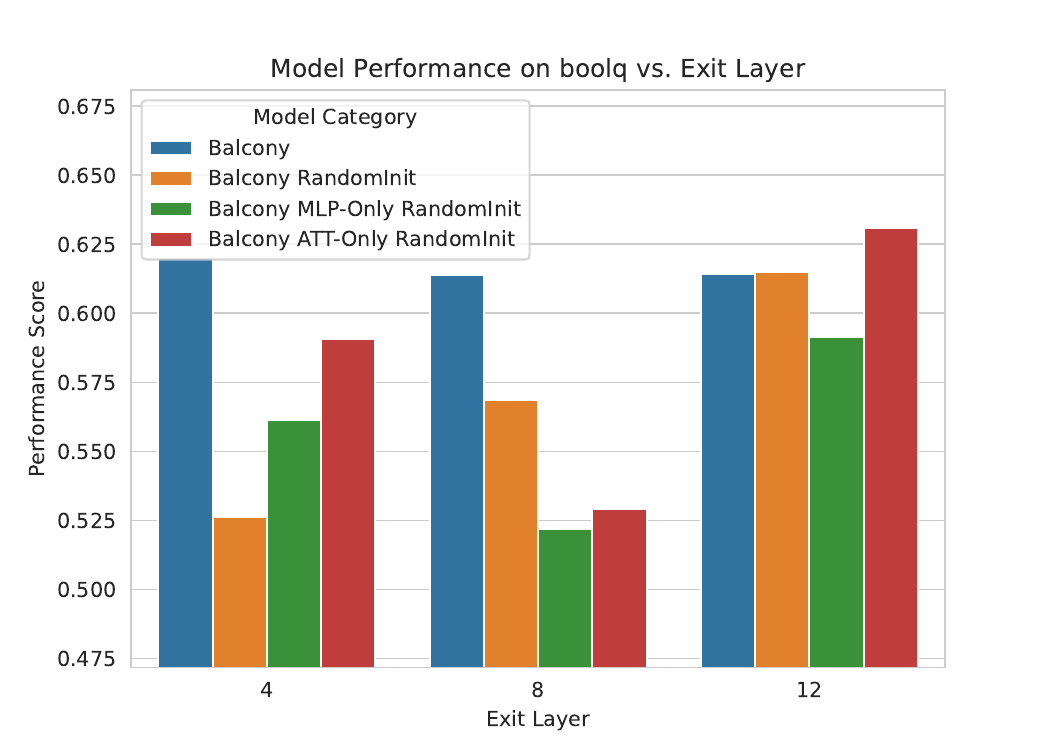}
    \end{minipage}
    \begin{minipage}{0.32\textwidth}
        \includegraphics[width=\linewidth]{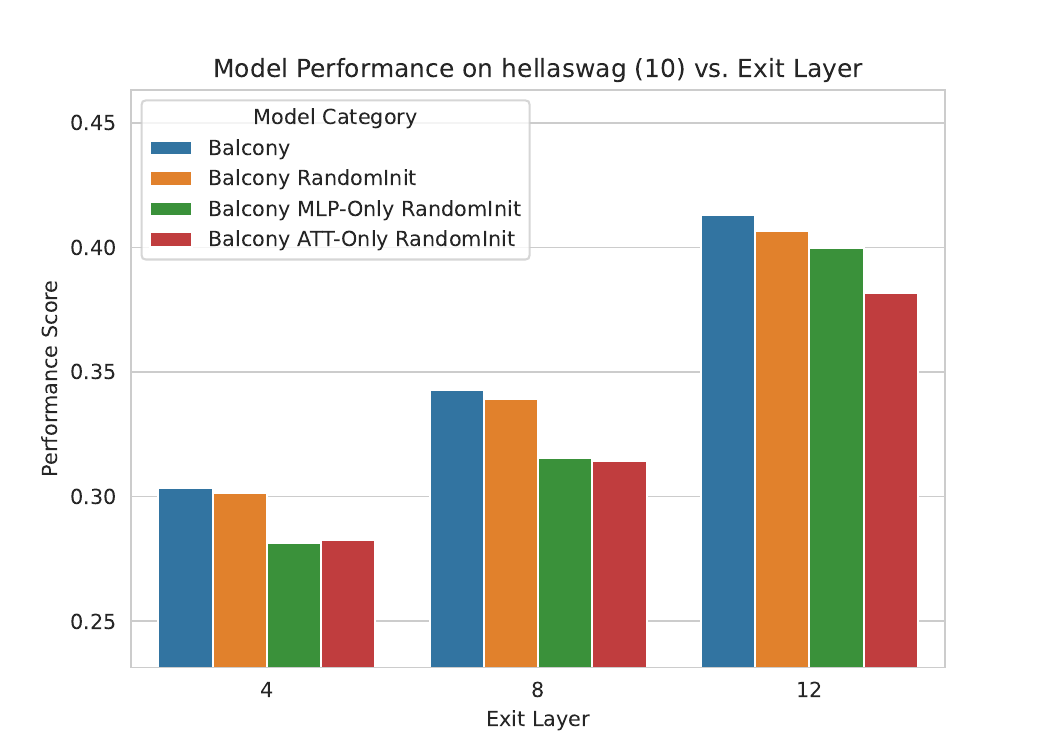}
    \end{minipage}
    \begin{minipage}{0.32\textwidth}
        \includegraphics[width=\linewidth]{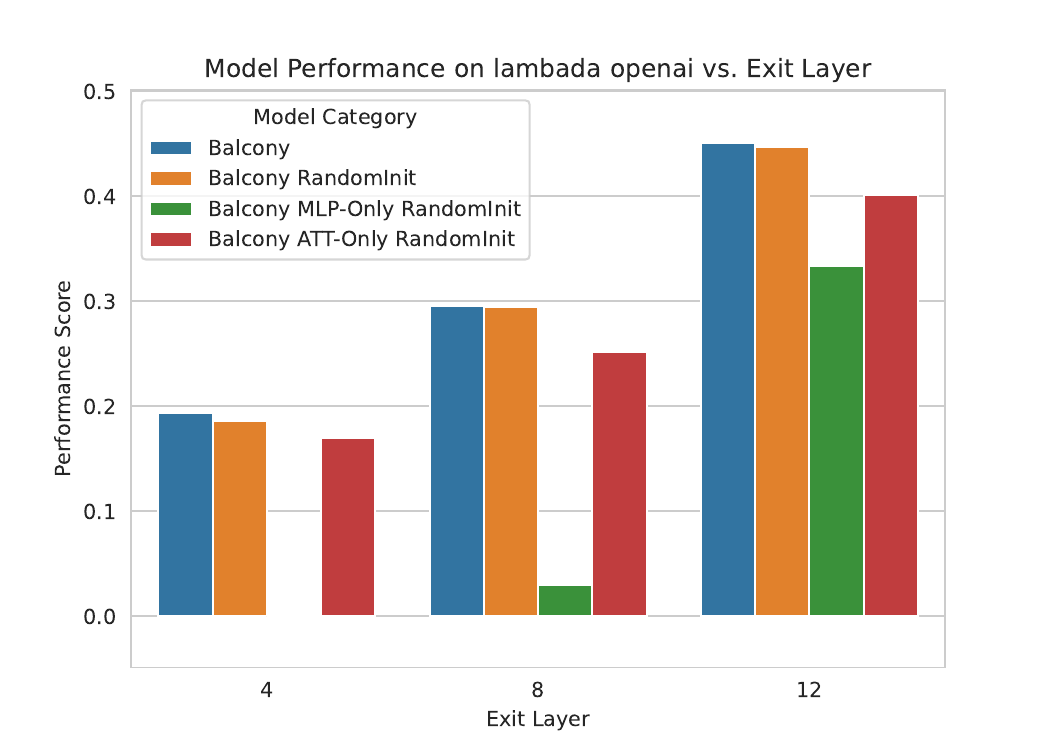}
    \end{minipage}
    \begin{minipage}{0.32\textwidth}
        \includegraphics[width=\linewidth]{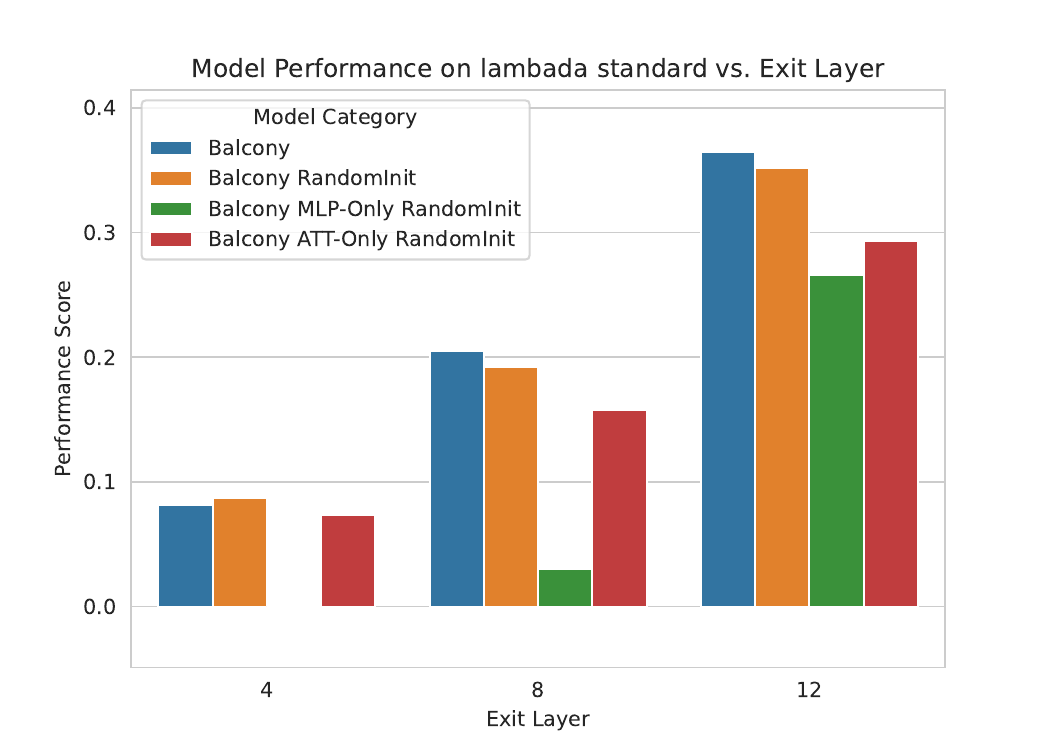}
    \end{minipage}
    \begin{minipage}{0.32\textwidth}
        \includegraphics[width=\linewidth]{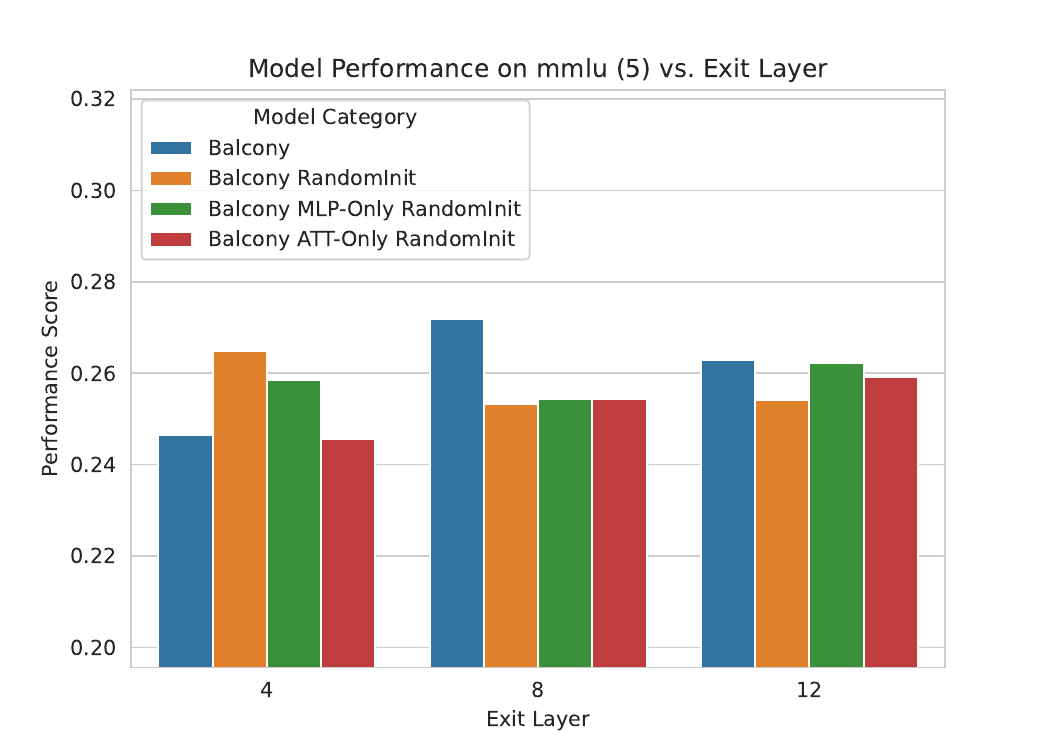}
    \end{minipage}
    \begin{minipage}{0.32\textwidth}
        \includegraphics[width=\linewidth]{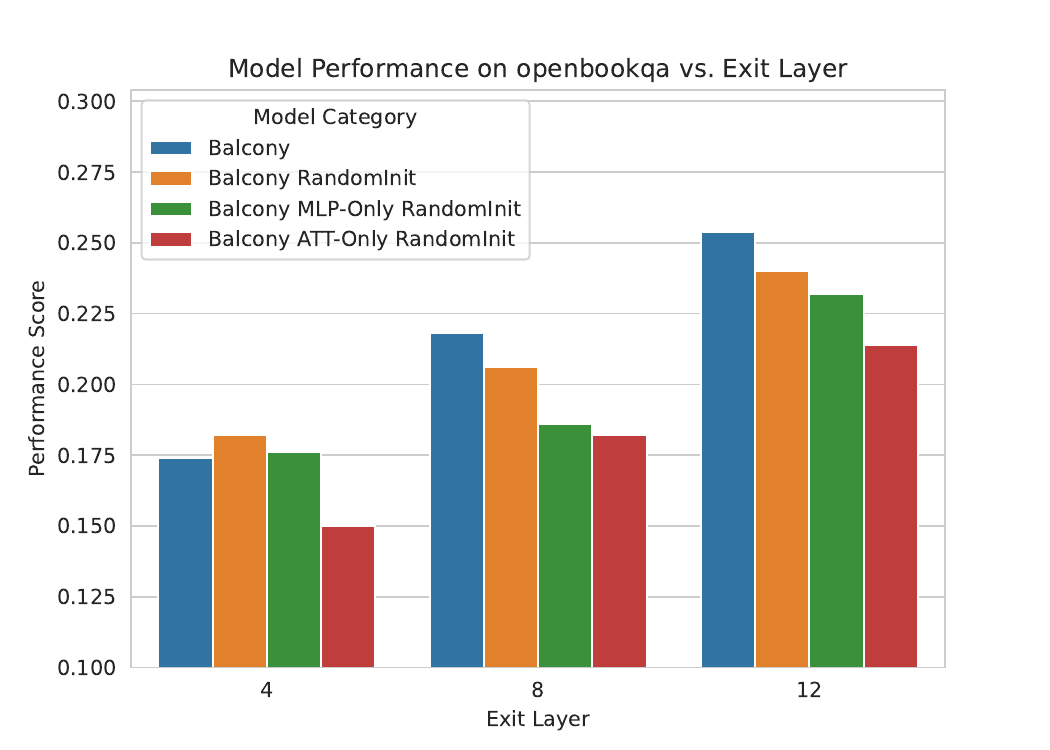}
    \end{minipage}
    \begin{minipage}{0.32\textwidth}
        \includegraphics[width=\linewidth]{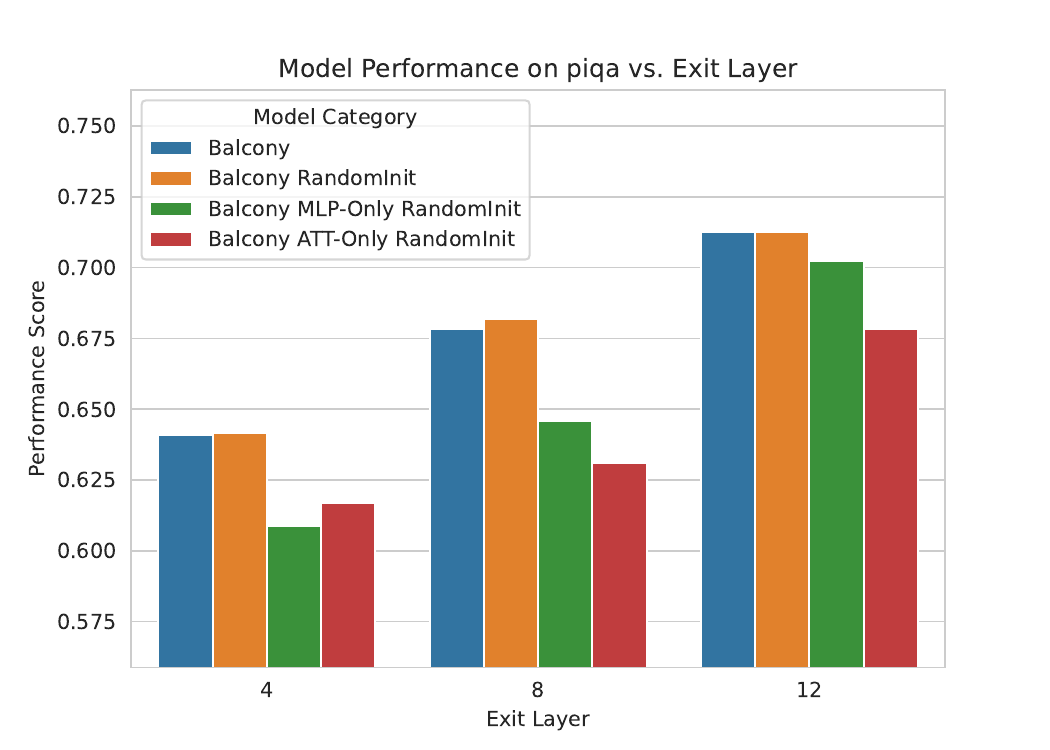}
    \end{minipage}
    \begin{minipage}{0.32\textwidth}
        \includegraphics[width=\linewidth]{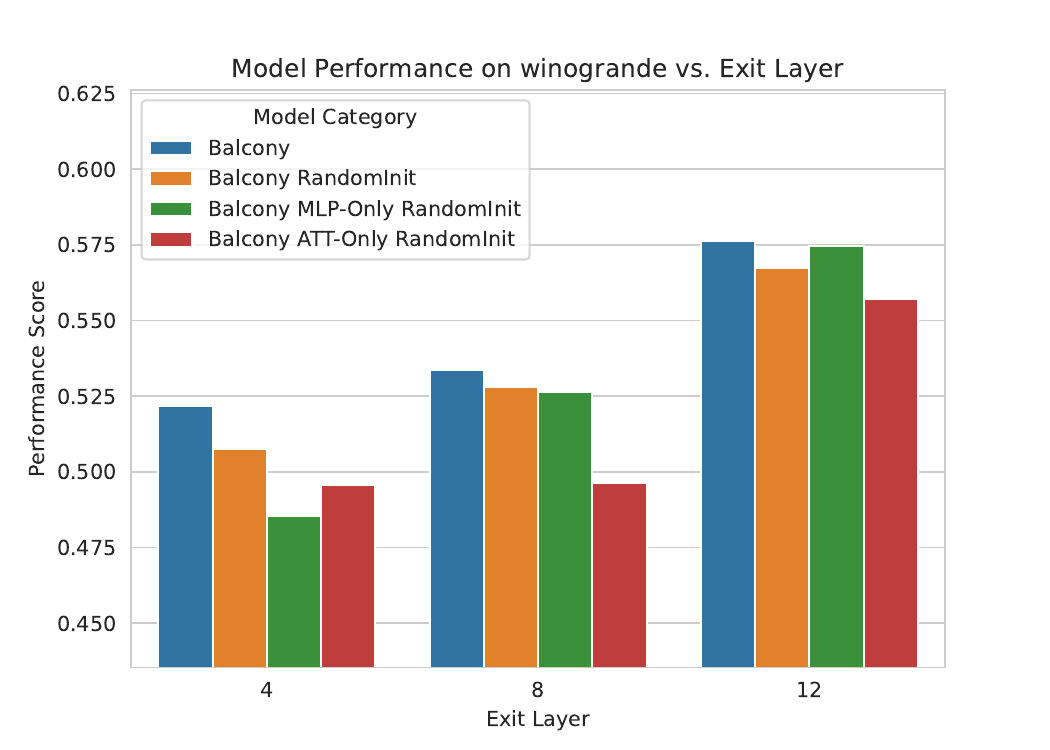}
    \end{minipage}
\end{figure*}

From our evaluations on eleven different benchmarks, we observe that our chosen initialization strategy, where the Balcony layer is initialized using the last layer of the full model, outperforms all other initialization variations. The results are shown in Figure~\ref{fig:mlpattn}

%\appendix

%\section{Appendix}
%\label{sec:appendix}

\end{document}